\documentclass[12pt]{article}
\usepackage{amsmath}
\usepackage{amsfonts}
\usepackage{times}
\usepackage{graphicx}
\graphicspath{{/}}
\usepackage{color}
\usepackage{multirow}

\usepackage{array}
\newcolumntype{C}[1]{>{\centering\arraybackslash}m{#1}}
\newcolumntype{L}[1]{>{\raggedright\arraybackslash}m{#1}}

\usepackage{float}
\usepackage{lscape}
\usepackage{subfloat}
\usepackage[round]{natbib}

\usepackage{rotating}
\usepackage{bbm}
\usepackage{latexsym}
\usepackage{subfig} 

\newcommand{\captionfonts}{\normalsize}

\makeatletter
\long\def\@makecaption#1#2{%
  \vskip\abovecaptionskip
  \sbox\@tempboxa{{\captionfonts #1: #2}}%
  \ifdim \wd\@tempboxa >\hsize
    {\captionfonts #1: #2\par}
  \else
    \hbox to\hsize{\hfil\box\@tempboxa\hfil}%
  \fi
  \vskip\belowcaptionskip}
\makeatother

\begin{document}

\hspace{13.9cm}1

\ \vspace{20mm}\\

{\LARGE Hardware-Amenable Structural Learning for Spike-based Pattern Classification using a Simple Model of Active Dendrites}

\ \\
{\bf Shaista Hussain}\\
\emph{School of Electrical and Electronic Engineering, Nanyang Technological University, Singapore.}\\
{\bf Shih-Chii Liu}\\
\emph{Institute of Neuroinformatics, University of Zurich and ETH Zurich, Switzerland.}\\
{\bf Arindam Basu}\\
\emph{arindam.basu@ntu.edu.sg}; Corresponding author\\
\emph{School of Electrical and Electronic Engineering, Nanyang Technological University, Singapore.}\\

%


\thispagestyle{empty}
\markboth{}{NC instructions}
\ \vspace{-0mm}\\
%
\begin{center} {\bf Abstract} \end{center}

This paper presents a spike-based model which employs neurons with functionally distinct dendritic compartments for classifying high dimensional binary patterns. The synaptic inputs arriving on each dendritic subunit are nonlinearly processed before being linearly integrated at the soma, giving the neuron a capacity to perform a large number of input-output mappings. 
The model utilizes sparse synaptic connectivity - where each synapse takes a binary value. The optimal connection pattern of a neuron is learned by using a simple hardware-friendly, margin enhancing learning algorithm inspired by the mechanism of structural plasticity in biological neurons. The learning algorithm groups correlated synaptic inputs on the same dendritic branch. Since the learning results in modified connection patterns, it can be incorporated into current event-based neuromorphic systems with little overhead. This work also presents a branch-specific spike-based version of this structural plasticity rule. The proposed model is evaluated on benchmark binary classification problems and its performance is compared against that achieved using Support Vector Machine (SVM) and Extreme Learning Machine (ELM) techniques. Our proposed method attains comparable performance while utilizing $10-50\%$ less computational resources than the other reported techniques.  


\section{Introduction}
Neuromorphic systems \citep{mead_neuromorphic} aim to emulate neuronal function by implementing the structure of the biological system. By using analog subthreshold Very Large Scale Integration (VLSI) circuits, these systems can implement a power-efficient model of neuronal processing. This approach is particularly useful for testing ideas about brain function that span across the levels of ion channels to neural micro-circuits and systems. There has already been significant effort in the past in this realm \citep{arthur_gamma,bar_ind_attention,iscas_bif_my,kai_neuron2,boahenliu2,orientation,misha4,hasler_marr}. The latest neuromorphic hardware systems are event-driven real-time implementations of energy-efficient, spike-based intelligent processing systems. Two key advantages of these systems are as follows: One, the event-driven way of processing leads to minimal power dissipation during long inactive periods \citep{dbn_mnist,liu_minitaur} and two, the systems are low-power because the transistors are operated in the sub-threshold region where currents are small. Usable event-driven sensors such as the Dynamic Vision Sensor retina \citep{tobi_imager, retina_2, retina_3} and the AEREAR2 cochlea \citep{liu_aerear} that produce asynchronous spike information about stimulus features in real-time environments are already available. There is a need to interface these sensors to event-based multi-neuron classifier systems that generate information only when necessary. These classifiers can benefit from the larger computational power of spiking neural networks \citep{maass_spiking_vc} and can also be used in novel applications related to brain-machine interfaces where the incoming spikes are generated by biological neurons \citep{bmi_review_nature}.

Reconfigurable spiking neural networks composed of neurons and synapses have been fabricated in VLSI. The networks are usually mixed hybrid analog-digital circuits on the chip while the connectivity information of the network is stored in an off-chip digital memory. An asynchronous communication protocol called the Address-Event Representation (AER) \citep{boahen_aer1,bernabe4,boahen_shi_hypercolumn,gert_vogel_ralph_jrnl} is typically used to route the neuronal spikes to and from the chip on a shared `fast' digital bus. Since the connectivity information is virtual, the user has full flexibility in the network reconfiguration. Synaptic weights are typically implemented using on-chip digital to analog converters (DAC) that are shared across synapses. Very few spike-based multi-neuron systems have reported solving real-world pattern classification or memorization problems comparable with state-of-the-art systems using algorithmic approaches. One major challenge in attaining a classification accuracy similar to a software implementation of a system on a computer is that statistical variations in VLSI devices reduce the accuracy of the synaptic weights. This problem is exacerbated by the fact that device mismatch is increasing with progressively shrinking transistor sizes. Hence, it has become increasingly important to design networks and algorithms which either exhibit robustness to variations or utilize low-resolution synapses which are better suited for robust implementation in VLSI systems. A spiking network classifier was implemented on a neuromorphic hardware system in which population coding was used to achieve tolerance against variability \citep{spike_class_insect}. This system however uses  a large number of synapses, which need to have high resolution weights to attain the performance reported in the study. This neuromorphic network achieved performance comparable with a standard machine learning linear classifier. However, nonlinear classifiers like the Support Vector Machine (SVM) can give far better results. 

In another study \citep{spike_class_rbm}, software simulation of a spiking neuromorphic system consisting of Restricted Boltzmann Machines (RBMs) was proposed with a learning rule based on Spike Timing Dependent Plasticity (STDP) for changing synaptic weights. These RBMs composed of leaky integrate and fire neurons were shown to be robust to the use of finite-precision synaptic weights with less than $3\%$ drop in performance for an MNIST recognition task when using $8$-bit vs $5$-bit precision weights. However, RBMs use a high number of synapses, almost two million recurrent connections, which makes its hardware implementation quite costly. In contrast to this, several studies have used binary synapses inspired from the evidence that biological synapses exist in only a small number of states \citep{syn_states_expt, binary_synapses_expt}. The use of binary synapses also provides a simple and robust solution to the variability due to device mismatch. A spike-based STDP learning algorithm using bistable synapses was implemented in \citet{fusi_mnist} with VLSI implementations in \citet{chicca_vlsispiking}, \citet{iscas_spikesynstdp} and \citet{mitra_spikeclassify}. This supervised learning algorithm was used to train a network to classify handwritten characters from the MNIST database \citep{fusi_mnist}. The classification accuracy was improved by having a pool of neurons that compute in parallel. However, this again leads to the problem of increased number of synapses by a factor equal to the number of neurons in the pool.

In this work, we present a bio-plausible structural learning rule for a neuron with nonlinear dendritic branches and suitable for implementation on neuromorphic chips. The paper is organized as follows: the full spike-based classification model of neurons with dendritic nonlinearity is presented in Section \ref{sec:arch}, followed by the reduced model for faster training (RM) in Section \ref{sec:red_model} and a margin-based reduced model (RMWM) in Section \ref{sec:modified}. The theoretical capacities of the RM and RMWM provide an upper bound on the number of patterns that can be correctly classified, as discussed in Section \ref{sec:theo_cap}. The learning rules for the RM and RMWM are based on the structural plasticity mechanism and are presented in Section \ref{sec:algo}. Further, a branch-specific spike-timing-based implementation of the structural learning rule (BSTDSP) is presented in Section \ref{sec:bstdsp_model}. The models were tested on a 2-class classification task (Sections \ref{sec:rm_results}-\ref{sec:results_bstdsp}) and the performance of the RMWM was compared against the SVM and ELM methods on a set of real-world binary databases in Section \ref{sec:results_uci}. The biological and computational relevance of our work is discussed in Section \ref{sec:discussion} followed by the conclusion and discussion of future work in Section \ref{sec:conclusion}.

\section{Model of Nonlinear Dendrites \& Structural Learning}
\label{sec:theory}

Various computational modeling studies have shown that neuronal cells having nonlinear dendrites (NLD) can perform complex classification tasks much more accurately than  their counterparts with linear subunits when the inputs are high-dimensional binary patterns \citep{mel_dendrite1,ijcnn_dendrite}. The presence of NLD provides a neuron with many nonlinear functions to better approximate the decision boundary of a complex classification problem. In this work we extend the NLD model of \citet{mel_dendrite1} to perform supervised classification of binary patterns consisting of rate encoded spike train inputs or patterns of single spikes.

\begin{figure}[!t]
\centerline{
\includegraphics[height=100mm,width=0.6\textwidth]{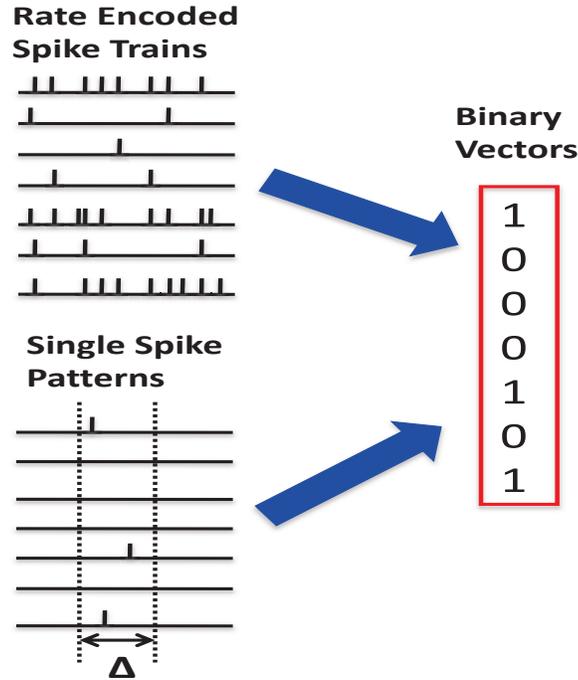}}
\caption{Conversion of binary patterns of spike inputs into binary vectors. An example of how rate encoded spike trains (top left) and single spikes (bottom left) are mapped to the same binary vector (right) is shown.}
\label{fig:input_conv}
\end{figure}

We consider two types of spike inputs - mean rate and single spike patterns. These spiking inputs encode binary patterns of numbers denoted as $\boldsymbol x \in R^d$ where each component $x_i \in \left\{0,1\right\}$. The first type is binary patterns of mean rate inputs which are defined such that for a spike train $s(t)=\sum_{t_k}{\delta(t-t_k)}$ arriving at an afferent, where $\delta(t)$ is a delta function and spike times $\left\{t_k\right\}\in [0, T]$\emph{\emph{}} ($T$ is the duration of the pattern), if the mean firing rate of the spike train (averaged over $T$) is higher than a threshold, the corresponding $x_i$ is set to $1$; else it is set to $0$ (Figure \ref{fig:input_conv}). The second type of spike inputs are binary patterns of single spikes in which $x_i=1$ if the $i^{th}$ afferent fires a single spike and $x_i=0$ if it fails to fire. Furthermore, all spikes arrive within a small time window ($\Delta$), i.e. $t_k \in [T_{syn}-\frac{\Delta}{2},T_{syn}+\frac{\Delta}{2}]$ where $T_{syn}$ is a fixed time between $[0, T]$. The binary inputs $\boldsymbol x$ are referred to as \textbf{binary vectors} in the rest of the paper. To reduce the training time, we have adopted the strategy of training the model on binary vectors $\boldsymbol x$ and testing on the original spike train patterns. We shall next describe the full spike-based model in Section \ref{sec:arch} followed by its reduced versions.

\begin{figure}[!t]
\centerline{\includegraphics[height=70mm,width=0.6\textwidth]{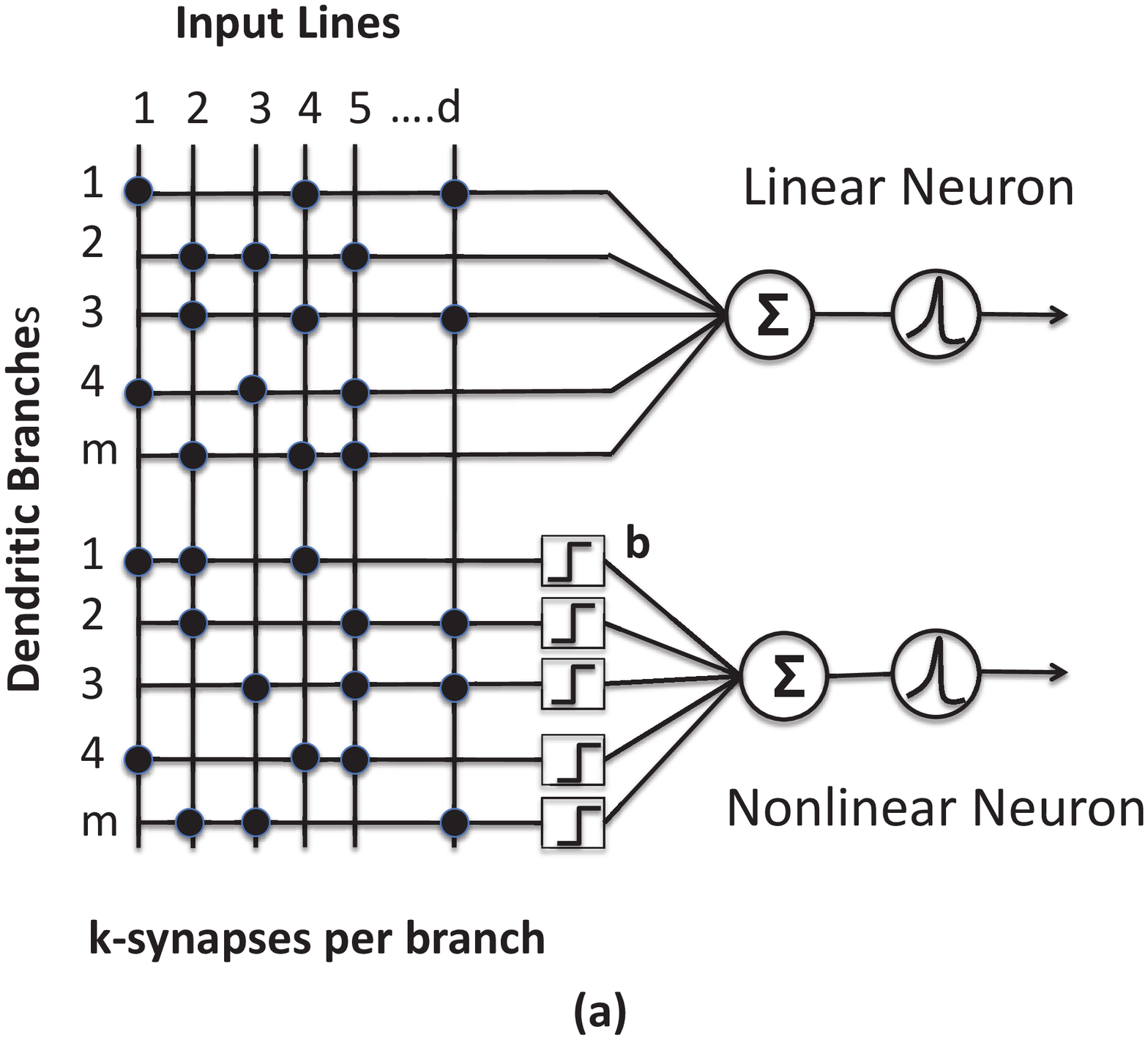}
\includegraphics[width=0.3\textwidth]{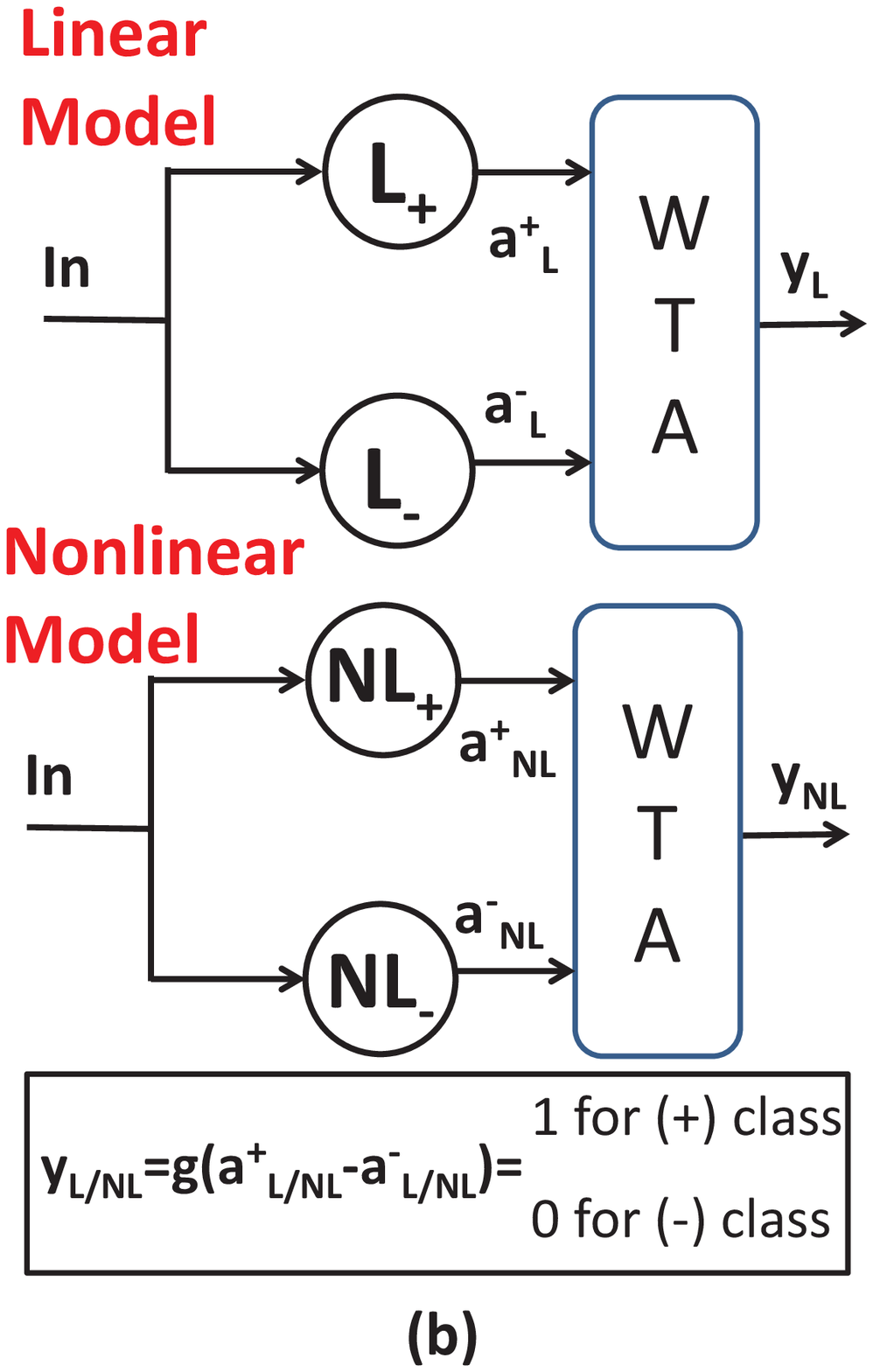}}
\caption{(a) The major difference between a traditional linearly summing neuron and our neuron with nonlinear dendrites is the lumped nonlinear function $b()$ for every dendritic branch. (b) Architecture of the 2-class pattern classifier that consists of a WTA to compare the outputs of two neurons trained to respond to patterns of classes (+) and (-) respectively by using the Heaviside function $g()$ to generate the output of the model.}
\label{fig:linvsNL}
\end{figure}

\subsection{Spike-Based Model}
\label{sec:arch}

We previously proposed a model of neuronal cells having a lumped dendritic nonlinearity in \citet{ijcnn_dendrite}. The nonlinear neuron (NL-neuron) consists of a simplified neuron with $m$ dendritic branches and $k$ excitatory synapses on each branch (Figure \ref{fig:linvsNL}(a)). Each synapse is driven by one of the $d$ components of the input. We enforce the constraint $k<<d$ in our model for several reasons: first, choosing the best `\emph{k}' connections out of `\emph{d}' possibilities is the basis of our learning rule; second, sparse connections reduce synaptic resources and third, this provides a means of regularization by reducing the number of free parameters. 

Let us consider the input spike train at the $i^{th}$ synapse of the $j^{th}$ branch given by:
\begin{equation}
s_{ij}(t)=\sum_{t^s_{ij}}\delta(t-t^s_{ij}) \label{eq:pre_train}
\end{equation}
where $t^s_{ij}$ denotes the times at which spikes arrive at this synapse. We use the leaky integrate and fire neuron model to generate the spike output using the following differential equation for the membrane voltage:
\begin{align}
\label{eq:lif}
&C\frac{dV_m(t)}{dt}+\frac{V_m(t)}{R}=I_{in}(t)\\
&\textrm{If } V_m(t)\geq V_{thr} \textrm{, } V_m(t) \to 0\textrm{;} \notag\\
&n_{spk} \to n_{spk}+1 \notag\
\end{align}
where $V_{thr}$ is the threshold for firing and $n_{spk}$ is the spike count, which is initialized to $0$ before every pattern presentation. The input current to the neuron, $I_{in}(t)$ is the sum of all dendritic currents, given by
\begin{align}
I_{in}(t)&=\sum_{j}I_{b,out}^j(t) \label{eq:Iin1}\\
I_{b,out}^j(t)&=b(I_{b,in}^j(t)) \label{eq:Iin2}\\
I_{b,in}^j(t)&=\sum_{i}w_{ij}(\sum_{t^s_{ij}<t}K(t-t^s_{ij})) \label{eq:Iin3}
\end{align}
Here $I_{b,out}^j(t)$ is the branch output current, $I_{b,in}^j(t)$ is the branch input current given by the sum of currents generated from all synapses on the $j^{th}$ branch, $w_{ij}$ is the weight of the $i^{th}$ synapse formed on the $j^{th}$ branch, $i\in \left\{1,\cdots d \right\}$, $j\in \left\{1,\cdots m \right\}$ such that $\sum_{i}w_{ij}=k$ where $w_{ij}\in Z_{+}$ which is the set of non-negative integers and $b$ denotes the nonlinear activation function of the dendritic branch. The postsynaptic current (PSC) generated by the input spikes at an afferent is given by $K(t-t^s_{ij})=K(t) \star s_{ij}(t)$ where the kernel function $K(t)$ \citep{tempotron} is given by:
\begin{equation}
\label{eq:kernel}
K(t) = I_0(exp(-t/\tau_f) - exp(-t/\tau_r))
\end{equation}
where $I_0$ is the normalization constant; and $\tau_f$ and $\tau_r$ are the fall and rise time constants of the PSC respectively. 

\begin{figure}[!t]
\centerline{
\includegraphics[height=55mm,width=0.6\textwidth]{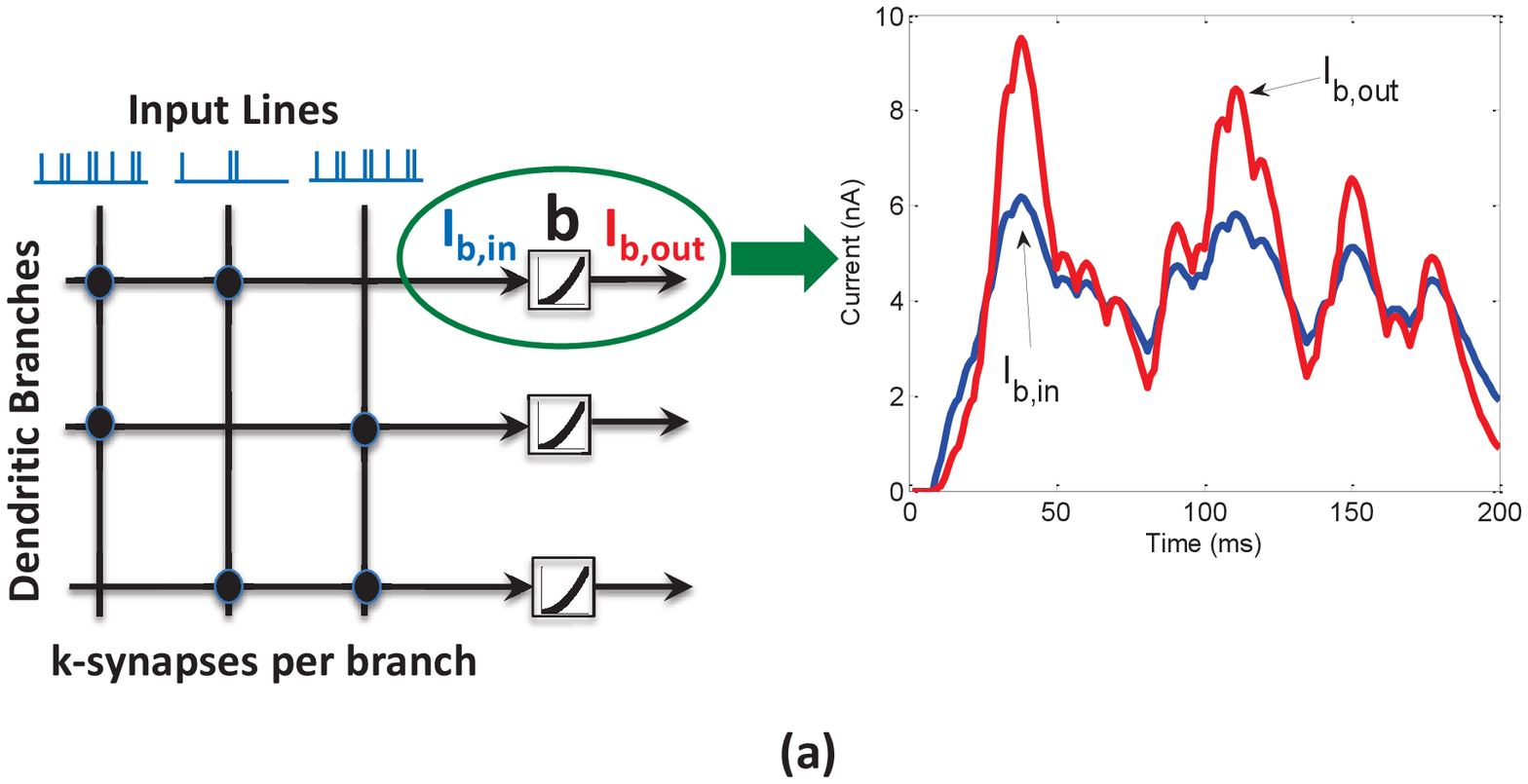}
\includegraphics[height=50mm,width=0.4\textwidth]{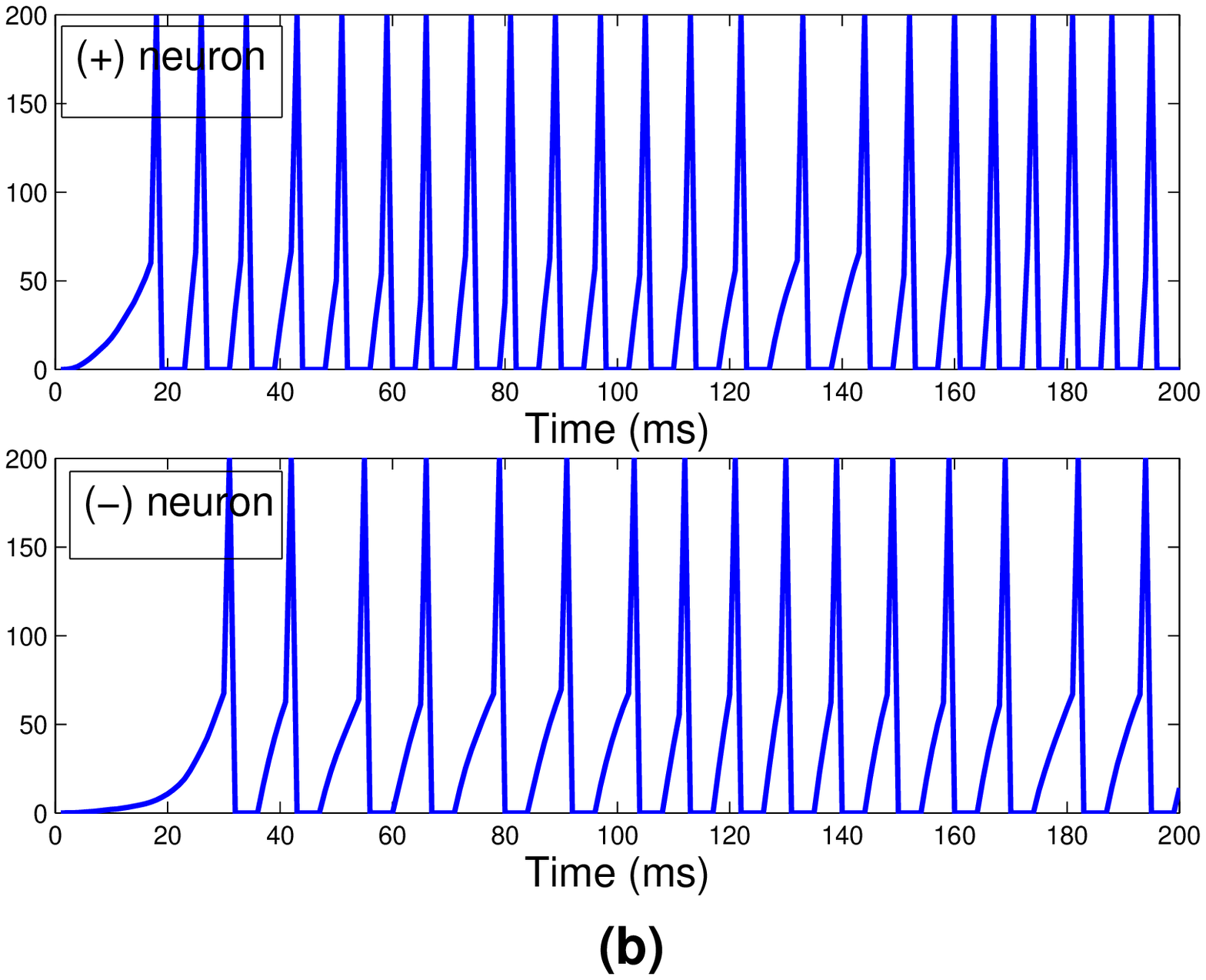}}
\caption{(a) The input and output of a dendrite (shown within ellipse) when a rate encoded spike pattern is presented. Time waveform of the input current, $I_{b,in}$ and the output current, $I_{b,out}$ of the branch are shown, where branch nonlinearity $b()$ is a quadratic function. (b) For an input pattern in class (+), the (+) neuron has a higher spiking rate. These output spike trains are input to a WTA which decides the winning neuron.}
\label{fig:sim_spike}
\end{figure}

We investigate the use of this model in a supervised classification of patterns belonging to two classes denoted as (+) and (-). Figure \ref{fig:linvsNL}(b) shows the classifier that uses a two-input winner take all (WTA) \citep{wta_lazarro,wta_chapter} circuit to compare the outputs of a pair of one excitatory (+) and one inhibitory (-) neuron trained to preferentially respond to patterns of class (+) or (-) respectively. The WTA circuit is used to select the winning neuron by generating the binary-valued output of the classifier, $y$ as follows:
\begin{align}
\label{eq:wta_sp}
\textrm{If } n_{spk}^+ > n_{spk}^- \textrm{, } y&=1 \notag\\
\textrm{else } y&=0
\end{align}

Figure \ref{fig:sim_spike}(a) shows the input and output of a dendrite (ellipse) when a rate encoded spike pattern is presented. Sample waveforms denote the time evolution of $I_{b,in}(t)$ and $I_{b,out}(t)$ for the dendrite for a case where $b()$ is a squaring function. Figure \ref{fig:sim_spike}(b) shows an example case where the firing rate of the neuron trained on mean rate encoded spike patterns of class (+) is higher for an input belonging to that class, i.e. $n_{spk}^+ > n_{spk}^-$. These output spike trains of the (+) and (-) neurons are used as inputs to the WTA which gives the classification output. 

\subsection{Reduced Model (RM)}
\label{sec:red_model}

For the two specific cases of spike trains considered here (mean rate inputs and single spike patterns), we can reduce the above model to reduce the time required for the training process. The reduced model was briefly introduced in \citet{ijcnn_dendrite} and its validity is discussed in Section \ref{sec:validity}. First, we present the details of the RM and then we describe the improved modified form proposed in this work. We first define the average activation level ($z_{syn,ij}$) at a synapse as:
\begin{equation}
z_{syn,ij}=\frac{1}{T}\int_{0}^{T}\sum_{t^s_{ij}}K(t-t^s_{ij})dt
\end{equation}
where $T$ is the duration of a pattern. Therefore, the average input current to a neuron is given by:
\begin{equation}
\tilde{I}_{in}=\sum_{j}b(\sum_{i}w_{ij}z_{syn,ij})
\end{equation}
Here $z_{syn,ij}$ is directly proportional to $x_{ij}$ which is the input arriving at the $i^{th}$ synapse on the $j^{th}$ branch, and can be written as $z_{syn,ij}=K_{syn}x_{ij}$. Therefore, the RM gives the average activation level of a NL-neuron as $a_{NL}=\hat{f}(\tilde{I}_{in})$, where $\hat{f}$ is the current-to-frequency function of the neuron (e.g. threshold linear functions can be used to approximate the expected firing rate of integrate and fire neurons \citep{gerst_book}). For simplicity, we have used a function with unity slope and zero threshold, leading to $\hat{f}(\tilde{I}_{in})=\tilde{I}_{in}$ (choosing other values of slope will not affect the result of classification where only a comparison is made between output of the two neurons). Therefore, the activation level $a_{NL}$ at the soma is given by:
\begin{equation}
\label{eq:neu_curr}
a_{NL}=\sum_{j}b(\sum_{i}w_{ij}z_{syn,ij})
\end{equation}
Similarly, for a linear neuron (L-neuron), the output of the soma is defined as $a_{L}=\sum_{j}\sum_{i}w_{ij}z_{syn,ij}$. In Figure \ref{fig:linvsNL}(a), the L-neuron is drawn in a manner similar to the NL-neuron to emphasize that the only difference between the two models is the absence or presence of the nonlinearity $b()$. Also, in all simulations both the linear and nonlinear neurons have the same total number of synapses (or synaptic resource), $s=m \times k$ so that we can compare performance of the two neuron types. Finally we can calculate the output of the classifier by modeling the WTA as an ideal comparator to get:
\begin{equation}
\label{eq:wta_op}
	y_{L/NL}=g(a_{L/NL}^{+}-a_{L/NL}^{-})
\end{equation}
where $g$ is a Heaviside step function that gives an output $1$ if the argument ($a_{L/NL}^{+}-a_{L/NL}^{-}$) is positive and $0$ otherwise. The subscript $L/NL$ of the variables denotes the corresponding output of a linear/nonlinear neuron. For convenience, we will drop this subscript in the remainder of the paper with the understanding that $y$ or $a$ refers to $y_{L}$/$y_{NL}$ or $a_{L}$/$a_{NL}$ in the context of linear/nonlinear neurons.

\begin{figure}[!t]
\centerline{
\includegraphics[width=0.6\textwidth]{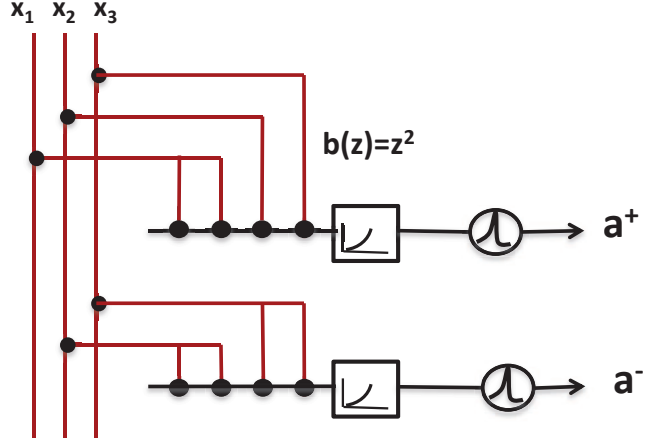}}
\caption{An example of (+) and (-) neurons with one dendrite each and a squaring nonlinearity. Connections on the branches with $4$ synapses each are shown for binary input vector $\boldsymbol x=[x_1$ $x_2$ $x_3]^T$.}
\label{fig:model_eg}
\end{figure}

The nonlinear dendritic subunits in the RM act as higher-order units which can capture nonlinear input-output relationships as in higher-order networks \citep{giles_1, sigma_pi_book, pi_sigma}. These nonlinear mappings can be realized by polynomial functions which are useful nonlinearities used to represent correlations among input components. We have used a squaring function for the dendritic nonlinearity, $b(z_j) = z_j^{2}/x_{thr}$, where $z_j=\sum_iw_{ij}z_{syn,ij}$ is the total synaptic activation on the $j^{th}$ branch. With this branch function, we can write the output of a NL-neuron as $a_{NL}=\frac{K^2_{syn}}{x_{thr}}\sum_{j}(\sum_{i}w_{ij}x_{ij})^2$. This nonlinearity is chosen for its easy implementation in hardware using current-mode circuits \citep{liu_indiveri_book,fpaa_jssc_my}. For our case of binary input vectors and binary synapses, setting $x_{thr} \geq 2$ ensures that at least two synapses on a NLD need to be active ($z_j \geq 2$) to evoke a supra-linear response. The use of a quadratic nonlinearity yields first and second-order terms in the decision function of the classifier. This is illustrated in the example shown in Figure \ref{fig:model_eg}, which shows (+) and (-) neurons with one dendrite each and input connections as shown. For the binary input vector $\boldsymbol x=[x_1$ $x_2$ $x_3]^T$ where $x_i \in {0,1}$, the two dendritic functions can be written as $(2x_1+x_2+x_3)^2$ and $(2x_2+2x_3)^2$, which generate cross-correlation terms $x_1x_2$, $x_2x_3$ and $x_1x_3$. Finally, the discriminant function $h(\boldsymbol x)$ can be written as the difference of the outputs of (+) and (-) neurons.
\begin{align}
h(\boldsymbol x)&=(2x_1+x_2+x_3)^2-(2x_2+2x_3)^2 \notag\\
&=4x_1^2-3x_2^2-3x_3^2+4x_1x_2+4x_1x_3-6x_2x_3 \notag\
\end{align}
where $h(\boldsymbol x)=0$ gives the decision boundary in the $3$-dimensional pattern space. If the dendritic nonlinearity is a polynomial of any order, $n$, it will generate terms like $x_i^px_j^q$ in each dendritic output function, where $p$ and $q$ are positive integers. Since for binary inputs $x_i^p=x_i$ and $x_i^px_j^q=x_ix_j$, therefore the connection matrix learned by the neurons (or equivalently the decision boundary) with any polynomial dendritic nonlinearity reflects the correlations present in the inputs. In other words, the (+) neuron should make connections reflecting those correlations present in patterns belonging to class (+) but not in class (-) and vice versa. This will be shown later in Section \ref{sec:train_rm}.

\begin{figure}[!t]
\centerline{
\includegraphics[width=0.6\textwidth]{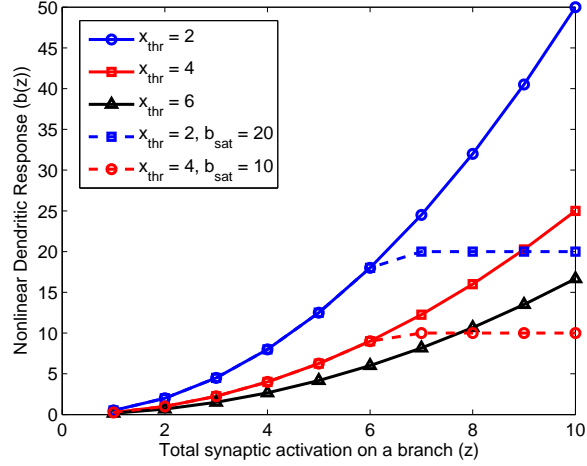}}
\caption{Dendritic nonlinear function $b(z)$ for different values of $x_{thr}$ and $b_{sat}$.}
\label{fig:b_NL}
\end{figure}

The use of a squaring nonlinearity for each dendritic branch will result in unrealistically large values for large inputs and large power dissipation in the corresponding analog VLSI hardware. Hence, we introduce a saturation level such that for $b(z_j)\geq b_{sat}$, $b(z_j)=b_{sat}$. Figure \ref{fig:b_NL} shows the nonlinear function $b(z)$ for different $x_{thr}$ values, with (dashed) and without (solid) the $b_{sat}$ term. This saturating dendritic activation function also conforms well with the electrophysiological measurements of synaptic summation within dendrites of pyramidal neurons \citep{mel_dendrite_b}. Fits using saturating polynomial functions on the experimental data measured in pyramidal neurons in the cited study is presented in Section \ref{sec:squareNL}.

\subsection{Reduced Model with Margin (RMWM)}
\label{sec:modified}

\begin{figure}[!t]
\centerline{\includegraphics[width=0.5\textwidth]{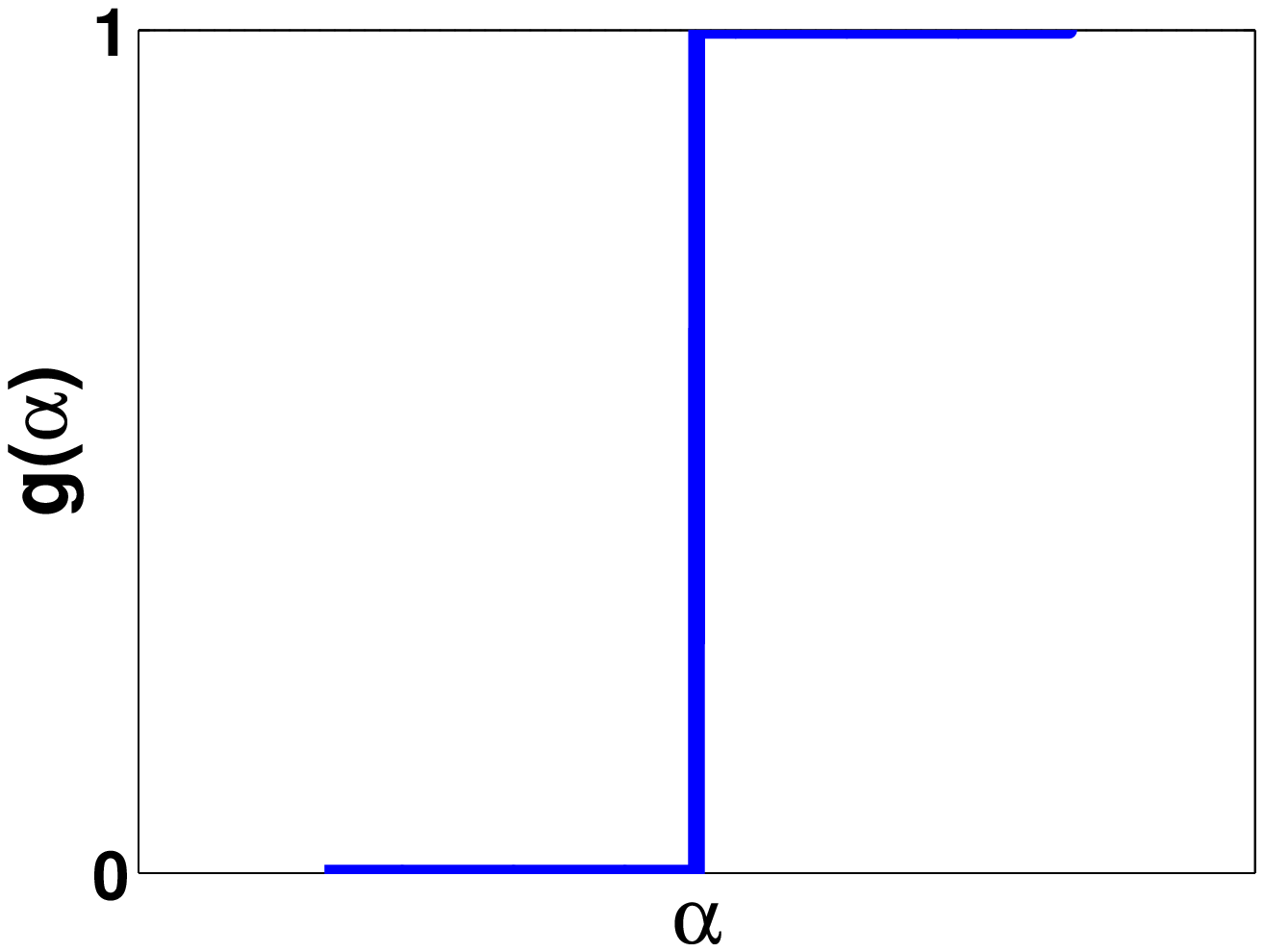}
\includegraphics[width=0.5\textwidth]{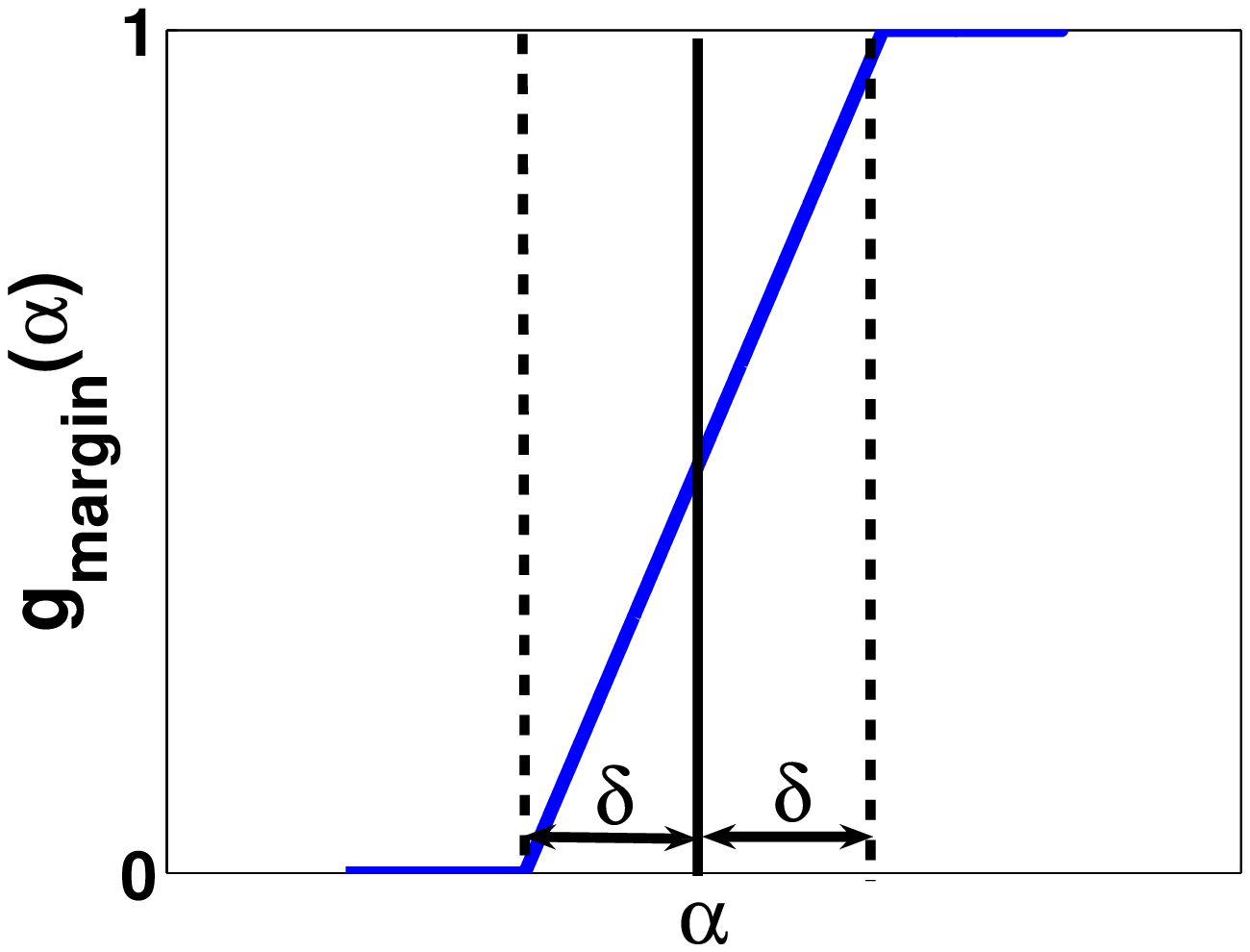}}
\caption{(a) Heaviside step function $g()$. (b) $g_{margin}$ function to enforce a margin $\delta$ around the classification boundary.}
\label{fig:g_fns}
\end{figure}

We have shown in \citet{ijcnn_dendrite} that when RM is tested with noisy spike inputs, the classification error is about $5$ times higher than the training error. Therefore, we describe modifications to this model with the objective of enhancing the robustness of the model to noisy inputs. This method uses a modified margin function $g_{margin}()$ function (Figure \ref{fig:g_fns}(b)) during training. We can define $g_{margin}():\Re\rightarrow\Re$, where $\Re$ is the set of real numbers, as given below:
\begin{align}
g_{margin}(\alpha)&=1 \textrm{ if }\alpha \geq \delta \label{eq:gnew}\notag\\
 &=0 \textrm{ if }\alpha \leq -\delta \notag\\
 &=\frac{0.5}{\delta}\alpha + 0.5 \textrm{ otherwise}
\end{align}	
where $\alpha$ denotes $a^+-a^-$. The Heaviside step function $g()$ (Figure \ref{fig:g_fns}(a)) is only used during the testing phase. This $g_{margin}()$ function enforces a margin $\delta$ around the classification boundary and as shown in Section \ref{sec:test_rmwm}, this margin improves the generalization performance similar to margin maximization in SVMs \citep{haykin_nn_book}. 

We found that the choice of $\delta$ depends on the size of the neuron measured by the number of branches and the number of synapses on each branch. The parameter $\delta$ can also be varied to achieve a trade-off between the training and the generalization performance, as shown in \citet{fusi_mnist}. A larger $\delta$ will give better generalization ability and reduce training performance while smaller values of $\delta$ will give optimal performance on the training set at the cost of reduced generalization performance. Therefore, an adaptive method was used to set the value of $\delta$ by assigning an initial high value $\delta_0$ and then reducing it gradually as discussed in Section \ref{sec:algo_modified}. 

Next, we propose a modification to the dendritic nonlinearity used in RM. This was motivated by the saturating nature of the $b()$ function and the fact that branch outputs can get saturated (Figure \ref{fig:b_NL}) leading to reduced classification performance of the model. This problem becomes more severe as the number of synaptic connections on each branch increases, which increases the average input to $b()$. Hence, we modify $b()$ so that a constant signal proportional to the mean synaptic activation of a branch is subtracted from the total excitatory synaptic activation of each branch. The constant signal is similar to the signal from the inhibitory population in the model proposed by \citet{fusi_mnist}. The new dendritic nonlinearity $b_{leak}()$ is defined as:
\begin{align}
b_{leak}(z_j)&=g(z_j-z_{leak,j})\frac{(z_j-z_{leak,j})^2}{x_{thr}} \textrm{ if }b_{leak}(z_j)<b_{sat} \label{eq:bleak}\notag\\
 &=b_{sat} \textrm{ otherwise} 
\end{align}	
where $g$ is the Heaviside step function and $z_{leak,j}$ is the average synaptic activation on the $j^{th}$ branch corresponding to the initial random connections, given by $z_{leak,j}=P(x_{i}=1)\times z_{syn,ij}\times k$, where $P(x_{i}=1)$ is the probability that $x_{i}=1$ and $k$ is the number of synapses on every branch of the neuron. Since, we consider a neuron with all the branches having same number of synapses, therefore, average activation on a branch before learning, $z_{leak,j}$ is same for all branches. Hence, we drop the branch subscript $j$ and denote $z_{leak,j}$ as $z_{leak}$ in the rest of the paper. Here, $z_{leak}$ can be regarded as an inhibitory signal to balance the excitation on each dendritic branch.

\subsection{Theoretical Capacity Predictions}
\label{sec:theo_cap}

For both the RM and RMWM, we predict the theoretical capacities of the L- and NL-neurons by estimating the number of distinct input-output functions that can be expressed by these neurons as a function of the total number of synapses, $s$, and the number of distinct input lines, $d$. We wanted to see if the added flexibility provided by the nonlinear dendrites allows us to relax the resolution of weights used. To do this, we use binary weights such that $w_{ij}=1$ if a connection exists and $0$ otherwise. Therefore, the capacity of these neurons can be given by the combinatorial expressions to count the number of input-output mappings, i.e. all possible ways in which input lines can connect to synapses on dendrites resulting in distinct memory fields. These expressions were derived in \citet{mel_dendrite1} and are provided here for reference. For a L-neuron, it is given by the number of ways in which $s$ synaptic sites can be drawn from $d$ input lines with replacement, equal to $\dbinom{s+d-1}{s}$. The spatial location of the $s$ synapses is not considered. For a NL-neuron, we first count the number of distinct branch functions available by drawing $k$ synapses from $d$ afferents with replacement, given by $f=\dbinom{k+d-1}{k}$ and then the number of ways in which $m$ branches of a NL-neuron can be drawn from $f$ possible functions, calculated as $\dbinom{f+m-1}{m}$, where total number of synapses, $s=m \times k$ is kept the same as that for L-neuron to allow direct comparison of capacities. The theoretical capacity calculated from these expressions is compared with the classification errors measured by training linear and nonlinear neurons on binary classification tasks as described in Section \ref{sec:train_rm}.

\subsection{Learning Algorithm}
\label{sec:algo}

In this section, we describe a learning rule which attempts to achieve classification performance in agreement with the theoretical capacity of a neuron, i.e. for a fixed number of synapses, minimum classification error is obtained when the predicted capacity is high. This learning rule is motivated by the feasibility of the equivalent hardware implementation. The goal of the learning process is to train the L- and NL-neurons so that the classifier output is $y=1$ and $y=0$ for input samples from the positive and negative classes respectively. A binary teacher signal, $o$ ($=1$ or $0$) guides the learning process by indicating the desired output during the training phase. The learning process can be regarded as a combinatorial search problem of selectively forming synaptic connections on each dendritic subunit. This learning rule is a supervised or error modulated form of Hebbian learning, which is based on the change in the connection strength of a synapse as a function of the presynaptic and postsynaptic activities. However, our learning algorithm involves the formation and elimination of synapses instead of weight modification, since we use binary weights.

\subsubsection{Learning Rule for RM}
\label{sec:algo_red}

To guide the learning process, a fitness function $c_{ij}$ is calculated for each synapse as the correlation between synaptic input and the output of the branch on which the synapse is located and modulated by the classification error. This correlation function is based on the fitness function used by \citet{mel_dendrite1}, which was derived from a stochastic gradient descent learning rule. Since, (+) and (-) neurons learn to respond preferentially to patterns belonging to class (+) and (-) respectively, the learning for the two neurons is complementary to each other. The fitness values for each pattern, $\Delta c_{ij}$ are computed using the following equations:
\begin{align}
\textrm{For (+) class neuron, for every pattern } \Delta c^+_{ij} &= x_{ij} b_{j} sgn(o-y) \label{eq:cell1}\\
\textrm{For (-) class neuron, for every pattern } \Delta c^-_{ij} &= -x_{ij} b_{j} sgn(o-y) \label{eq:cell2}
\end{align}
where $b_j$ denotes the output of the $j^{th}$ branch given by $b_j=b(z_j)$ and $sgn()$ is the signum function with a value of $1$ for $o>y$, $-1$ for $o<y$ and $0$ for $o=y$. Note that the error term $sgn(o-y)$ only modulates the sign of $\Delta c_{ij}$. The $c_{ij}$ values are then calculated over an epoch as:
\begin{align}
c^+_{ij} &= <\Delta c^+_{ij}> \label{eq:cell1avg}\\
c^-_{ij} &= <\Delta c^-_{ij}>  \label{eq:cell2avg}
\end{align}
where $<.>$ indicates averaging over the entire training set.

The error term ensures that only wrongly classified patterns contribute to $c_{ij}$. This implies that the connection changes based on the input correlations, have minimal effect on the patterns already learned, which agrees with the basic adaptive learning principle of \emph{minimal disturbance} \citep{widrow_review}. The connection changes are done by using the following logic: a synapse with the smallest $c_{ij}$ value corresponds to a poorly-performing synapse and is a candidate for replacement. Since we operate with fixed synaptic resources, this synapse needs to be replaced by another one. To do this, a set of silent synapses are first formed on the chosen branch as possible candidates for the new connection. The silent synapses are randomly selected from the set of input lines in addition to the existing synapses and do not contribute to the dendritic activity. The silent synapse with the highest $c_{ij}$ is chosen as the replacement. Hence, we modify the grouping of connections on a dendrite guided by a Hebbian-like learning rule. This type of synaptic clustering has been observed experimentally in \citet{act_dep_clust} showing activity-dependent clustering of synaptic inputs on developing hippocampal dendrites. Moreover, our calculation of $c_{ij}$ can be implemented easily in hardware since it needs only local information while the global error signal is a binary value.

The input connections for a neuron with $m$ branches and $k$ synaptic contacts per branch were initialized by randomly selecting afferents from one of the $d$ input lines with weight $w=1$. We trained our model on $P$ input pattern vectors ($\boldsymbol x$) belonging to (+) and (-) classes. The learning process comprises the following steps in every iteration:
\begin{itemize}
\item[(1)] The activation of each dendrite, $z_j$ is determined and then the sum of all branch responses is calculated as in equation \eqref{eq:neu_curr} for both (+) and (-) neurons, giving $a^{+}(\boldsymbol x)$ and $a^{-}(\boldsymbol x)$ for all $P$ patterns. The overall output of the classifier is computed according to equation (\ref{eq:wta_op}).
\item[(2)] The error is calculated by averaging the absolute error $\lvert o-y\rvert$ over the entire batch of $P$ patterns, which we describe as mean absolute error (MAE). Since in our case, $o$ and $y$ are binary-valued, the absolute error is always $1$ for every misclassified pattern and hence, the MAE is equal to the fraction of misclassified patterns.
\item[(3)] A random set $T$ consisting of $n_{T}$ ($<s$) synapses is selected as candidates for replacement.
\item[(4)] The synapse with the lowest $c_{ij}$ in $T$ is replaced by the best-performing (maximum $c_{ij}$) synapse in a random replacement set $R$ consisting of $n_{R}$ ($<d$) synapses. The set $R$ was created by placing $n_{R}$ `silent' synapses from $d$ input lines on the branch with the lowest $c_{ij}$ synapse. They do not contribute to the calculation in step (1).
\item[(5)] Synaptic connections are modified if the replacement led to either a reduction or no change in MAE. If the MAE increased with the replacement, a new replacement set $R$ is created. If the MAE does not decrease after repeating this step $n_{ch}$ ($=100$) times, we assume a local minimum is encountered. We then do a replacement with the last choice of $R$, even if it increases MAE in an attempt to escape the local minimum. The connection matrix corresponding to the local minimum that gave the least error is saved.
\item[(6)] Steps (1) to (5) are repeated until either all the patterns have been memorized or $n_{min}$ ($=100$) local minima are encountered. At this point, the learning process is stopped. The connection patterns corresponding to the best minimum become the final learned synaptic connections of the neuron.
\end{itemize}

The fraction of synapses to be replaced in each iteration is similar to a learning rate parameter in the traditional weight modification algorithm. A minimum of $1$ out of randomly selected $n_T$ out of all $s$ synapses is replaced in each iteration for smooth convergence to ideal topologies. The largest possible size ($n_R$) of the replacement set is equal to $d$, the dimension of the input. Using $n_R=d$ should give better optimization results but this increases computational cost. Therefore we chose $n_R$ as a fraction of $d$ which is a good trade-off between the computational cost and quality of the solution to the optimization problem. Finally, the RM trained on binary input vectors is tested on spiking inputs using the equivalent spike-based model and the connection matrix after training. Here, we wanted to test how well our reduced abstract learning rule can transfer the learned performance to classify the original spike patterns or the noisy spike inputs when the temporal dynamics of a spike-based model are introduced.

\subsubsection{Learning Rule for RMWM}
\label{sec:algo_modified}

The learning algorithm for the RMWM is the same except that the fitness values are calculated as:
\begin{align}
\textrm{For (+) class neuron, for every pattern } \Delta c^+_{ij} &= x_{ij} b_{leak,j} sgn(o-y) \label{eq:cell1_mod}\\
\textrm{For (-) class neuron, for every pattern } \Delta c^-_{ij} &= -x_{ij} b_{leak,j} sgn(o-y) \label{eq:cell2_mod}
\end{align}
where, $y=g_{margin}(\alpha)$. The fitness values $c_{ij}$ are calculated by taking the average of $\Delta c_{ij}$ for all patterns as shown in equations \eqref{eq:cell1avg} and \eqref{eq:cell2avg}. The value of margin $\delta$ for the $g_{margin}()$ function as given in equation \eqref{eq:gnew} is determined using the following steps:

\begin{itemize}
\item[(1)] Training is first performed using the \emph{$g()$ function} and then the corresponding spike inputs are presented to the model.
\item[(2)] The values of $\alpha$ are recorded for all cases in which the network misclassified the spiking inputs. 
\item[(3)] A margin of $\delta_0$ given by the highest value of $\alpha$ for a wrongly classified pattern is introduced around the classification boundary to reduce the misclassified cases. 
\item[(4)] $\delta$ value is reduced to $80\%$ of $\delta_0$ if the learning algorithm gets stuck in the same local minimum for $5$ consecutive times and is reduced to $80\%$ of its present value everytime this condition is encountered.
\end{itemize}
After the completion of training, the testing is done in the same way as in the case of RM, on spike inputs using the spike-based model.

\subsection{Spike-Time Based Learning}
\label{sec:bstdsp_model}
Our abstract models (RM and RMWM) have used the $c_{ij}$ values to guide structural plasticity; however, no specific spike time based mechanism to achieve this computation has been shown so far. In other words, all learning is done on an abstract model while only the testing has involved spike trains. We present in this section a possible spike time based rule that enables online calculation of $c_{ij}$ using an anti-Hebbian rule for excitatory synapses. Such ``reverse STDP" (RSTDP) rules have also been widely reported in biological measurements \citep{rstdp1, rstdp2, rstdp3}. Our spike-based learning rule is inspired from a recent study in which RSTDP in concert with hyperpolarization of the postsynaptic neuron or spike-frequency adaptation was used to realize variants of the perceptron learning rule \citep{NIPS2013_5054, spike_freq_perc} by modifying synaptic weights. The algorithm enables associative learning where a teacher signal is present during the learning process. In our case, we use a similar concept but in contrast to \citet{NIPS2013_5054}, we do not modify the weights but instead calculate the change in $c_{ij}$. Another major difference in our work compared to \citet{NIPS2013_5054} is that the learning rule for a synapse is modulated by its dendritic branch activation level and is hence termed branch-specific. This concept of branch specific plasticity has also been observed to a certain degree in biology \citep{dend_coupling, dend_coupling2} and has been used to model feature binding in a single neuron \citep{maass_bsp}. Finally, there is a question of what physical variable in biology might represent $c_{ij}$? There is evidence that calcium concentration in spines is a correlation sensitive, spike-time dependent signal and has been implicated as a guide for structural plasticity \citep{cij_calcium}. We compute the correlations in a branch-specific manner as discussed next in our branch-specific spike time dependent structural plasticity (BSTDSP) model.  

\subsubsection{BSTDSP: Model Details}

The basic architecture is similar to the nonlinear model shown in Figure \ref{fig:linvsNL}(b) consisting of (+) and (-) neurons, and the outputs of the neurons connected to a WTA circuit. The dynamics of the membrane potential $V_m(t)$ is as follows:
\begin{align}
\tau_V\frac{dV_m}{dt}&=(u-V_m)+I_{in}(t) \label{eq:lif_bstdsp}\\
\tau_u\frac{du}{dt}&=-u\\
\textrm{If } V_m\geq V_{thr} \textrm{, } &V_m \to V_{reset}\textrm{;} \notag\\
&u \to u_{reset}\textrm \notag\\
u_{reset} &= V_{reset} < 0 \notag\\
&n_{spk} \to n_{spk}+1 \notag\
\end{align}
where $u$ denotes a hyperpolarization variable which relaxes back to $0$ with a time constant $\tau_u$ and is set to $u_{reset}$ after a postsynaptic spike; $I_{in}(t)$ is the input current given by equations \eqref{eq:Iin1}-\eqref{eq:Iin3}; and $\tau_V$ and $\tau_u$ are the time constants governing the fast and slow dynamics of the membrane voltage and hyperpolarization respectively. Note that the differential equation for $V_m(t)$ in equation \ref{eq:lif} only had one state variable while equation \ref{eq:lif_bstdsp} has two. This allows us to separate the time scale for synaptic current based depolarization and post spike hyperpolarization i.e. recovery from hyperpolarization, which is much slower. It will be shown later in Section \ref{sec:learning} that this hyperpolarization allows us to build a margin for classification. The output of the model is computed as: 
\begin{equation}
y=g(n_{spk}^+ - n_{spk}^-)
\end{equation}

The presynaptic spike train $s_{ij}(t)$ is given by equation \eqref{eq:pre_train} and the postsynaptic spike train is denoted by:
\begin{equation}
r(t)=\sum_{t_{post}}\delta(t-t_{post})
\end{equation}
where $t_{post}$ is the postsynaptic spike time. These pre- and postsynaptic spikes also drive exponentially decaying memory traces--the presynaptic trace $\bar{s}$ and the postsynaptic trace $\bar{r}$ given by:
\begin{align}
\bar{s}&=exp(-t/\tau_{pre}) \\
\bar{r}&=exp(-t/\tau_{post})
\end{align}

\subsubsection{BSTDSP: Learning Rule and Parameter Choice}
\label{sec:learning}
To understand the learning process intuitively, let us consider the case of 2-class synchronous single spike inputs such that each afferent either fails to fire or fires a spike at a fixed time $T_{syn}$. During learning for $o=1$, a teacher signal forcing a postsynaptic spike (at time $t=1$ ms) is present for the (+) neuron and absent for the (-) neuron. For $o=0$, vice versa conditions for the teacher signal exist.

Our structural plasticity rule used to calculate the correlation value $c_{ij}$, is analogous to the RSTDP rule such that if a postsynaptic spike arrives before a presynaptic spike, then the correlation value for that synapse is increased. The condition for decreasing correlations does not use postsynaptic spike--instead, the relevant event is when the membrane voltage $V_m(t)$ crosses a subthreshold voltage $V_{st}$ from below, where $0<V_{st}<V_{thr}$. These subthreshold crossing events occurring at times $t_{st}$ are denoted by $r_{st}(t)=\sum_{t_{st}}\delta(t-t_{st})$. Depression happens when presynaptic spike time $t^s_{ij}$ occurs before $t_{st}$. The reason for using $r_{st}(t)$ instead of $r(t)$ is to enforce a margin as will be explained later in this section. These plasticity conditions are used in a branch-specific manner to compute the $\Delta c_{ij}(t)$ values for each pattern presentation, which are utilized to change connections on each branch. The learning rule can be written as:
\begin{align}
\textrm{For (+) class neuron, for every pattern } \Delta c^+_{ij}(t)&=I_{b,out}^{j+}(t) \bar{r}^{ +}(t) s^+_{ij}(t) - \gamma I_{b,out}^{j+}(t) \bar{s}^+_{ij}(t) r^+_{st}(t) \\
\textrm{For (-) class neuron, for every pattern } \Delta c^-_{ij}(t)&=I_{b,out}^{j-}(t) \bar{r}^-(t) s^-_{ij}(t) - \gamma I_{b,out}^{j-}(t) \bar{s}^-_{ij}(t) r^-_{st}(t)
\end{align}
where $\gamma$ is a constant used to balance potentiation and depression and ensure that $c_{ij}=0$ when a pattern is learned. The term $I_{b,out}^j(t)$ (from equation \eqref{eq:Iin2}) indicates branch specificity and is one major difference from \citep{NIPS2013_5054}. The correlation values averaged over all the patterns can be written as: $c^+_{ij} = <\Delta c^+_{ij}(t)>$ and $c^-_{ij} = <\Delta c^-_{ij}(t)>$. The connection changes based on $c_{ij}$ values are done as discussed in equations \eqref{eq:cell1} and \eqref{eq:cell2} in Section \ref{sec:algo_red}.

\begin{figure}
\begin{center}
\subfloat[]{\includegraphics[width=0.5\textwidth, height=0.25\textheight]{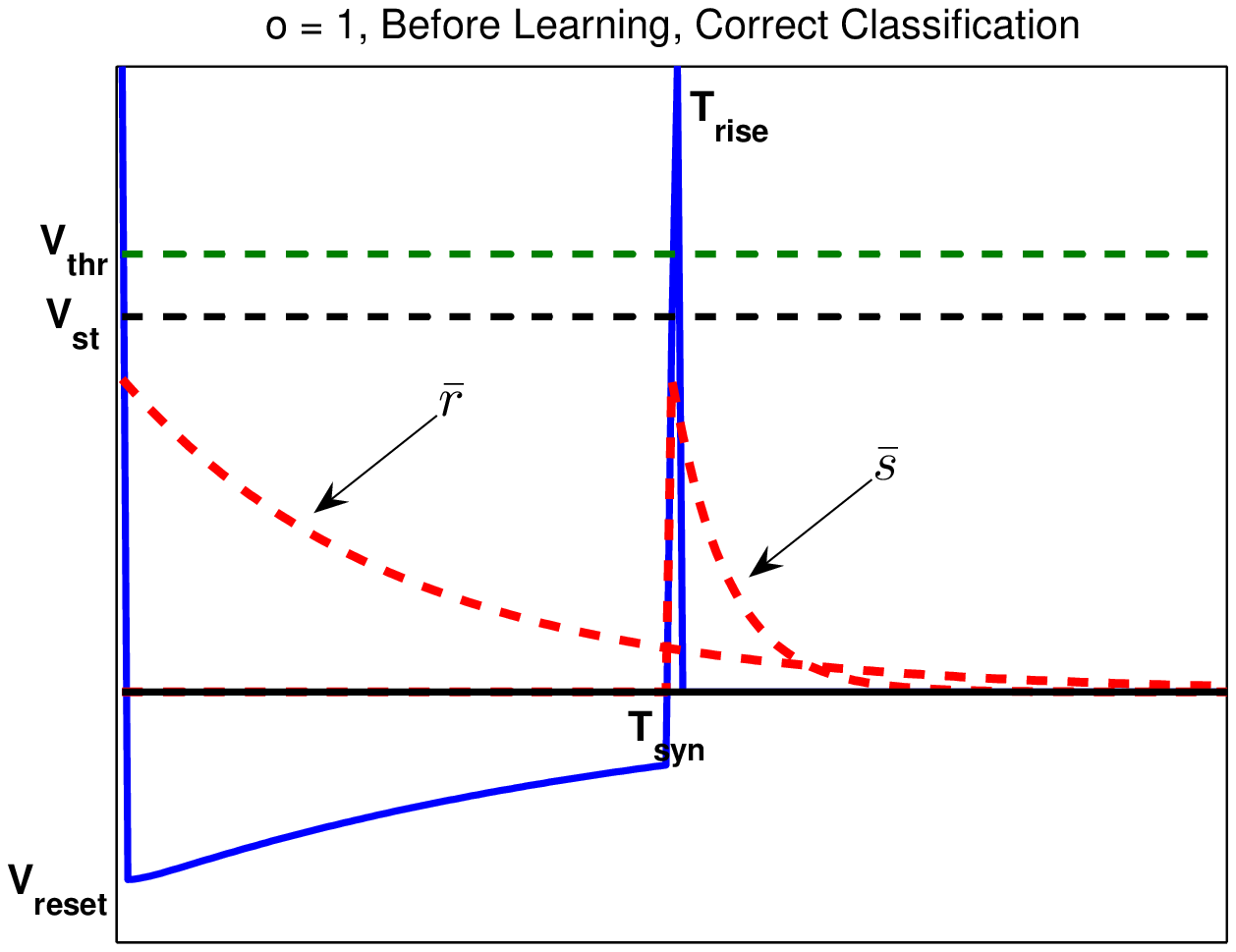}} 
\subfloat[]{\includegraphics[width=0.5\textwidth,height=0.25\textheight]{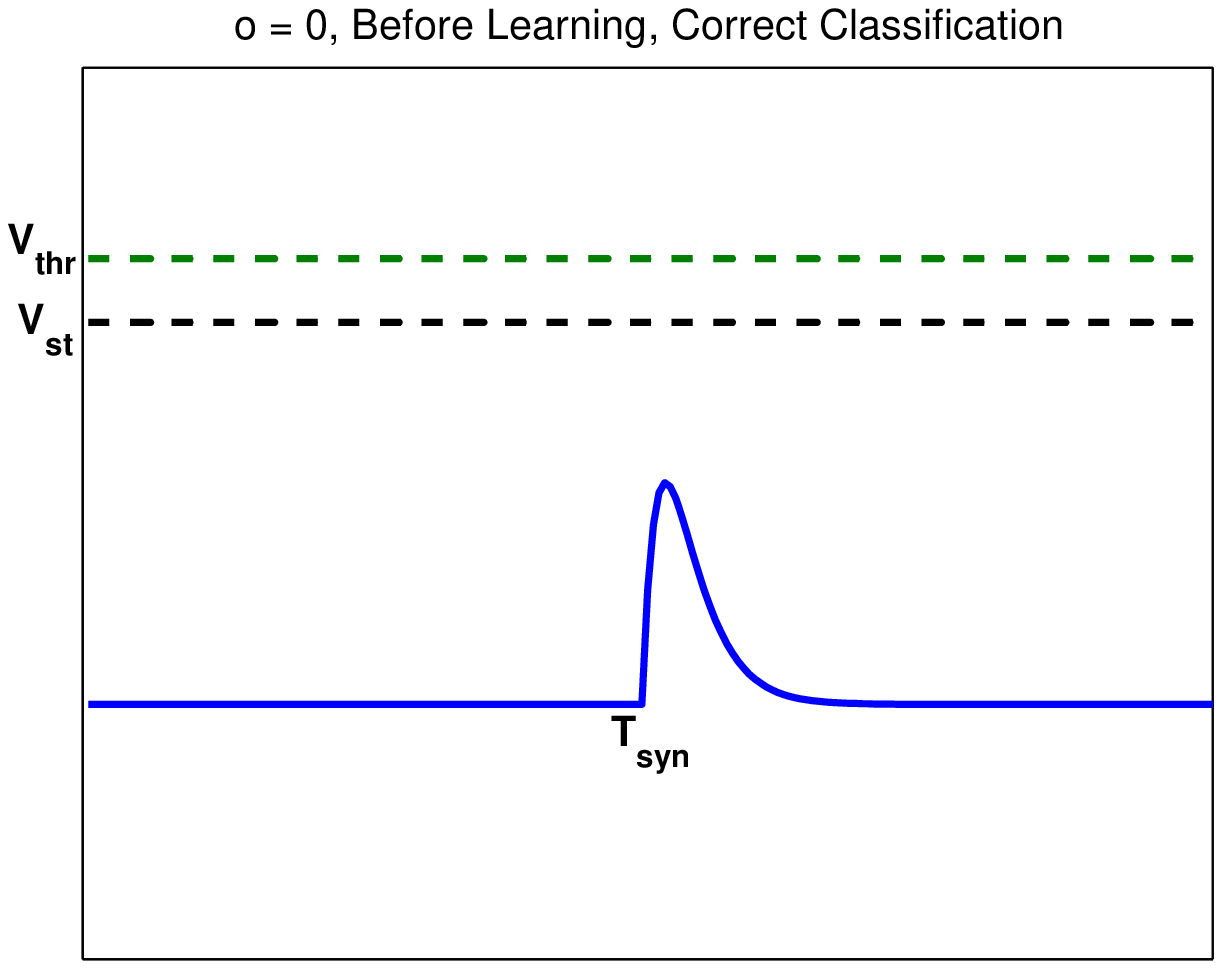}}
\\
\noindent 
\subfloat[]{\includegraphics[width=0.5\textwidth, height=0.25\textheight]{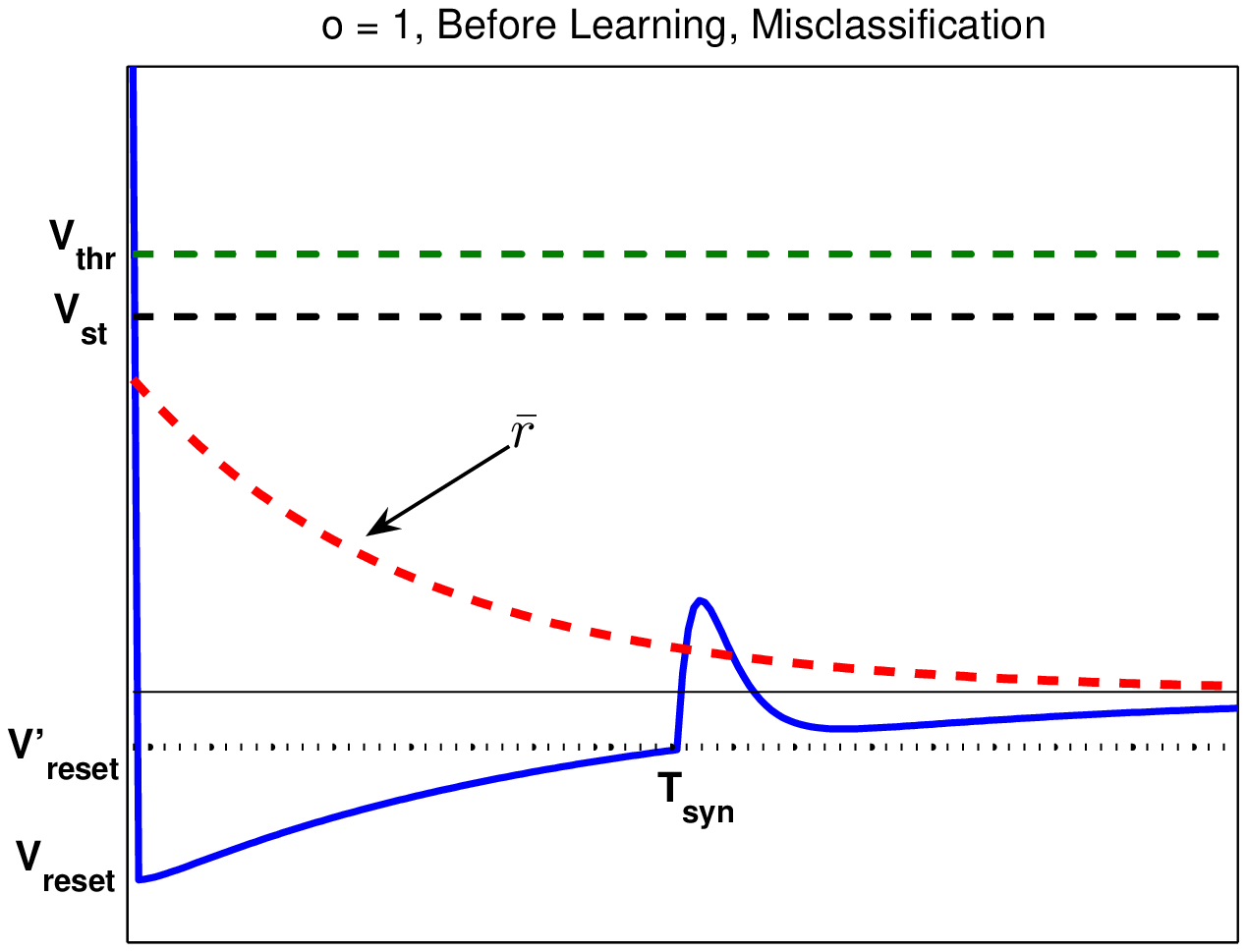}} 
\subfloat[]{\includegraphics[width=0.5\textwidth,height=0.25\textheight]{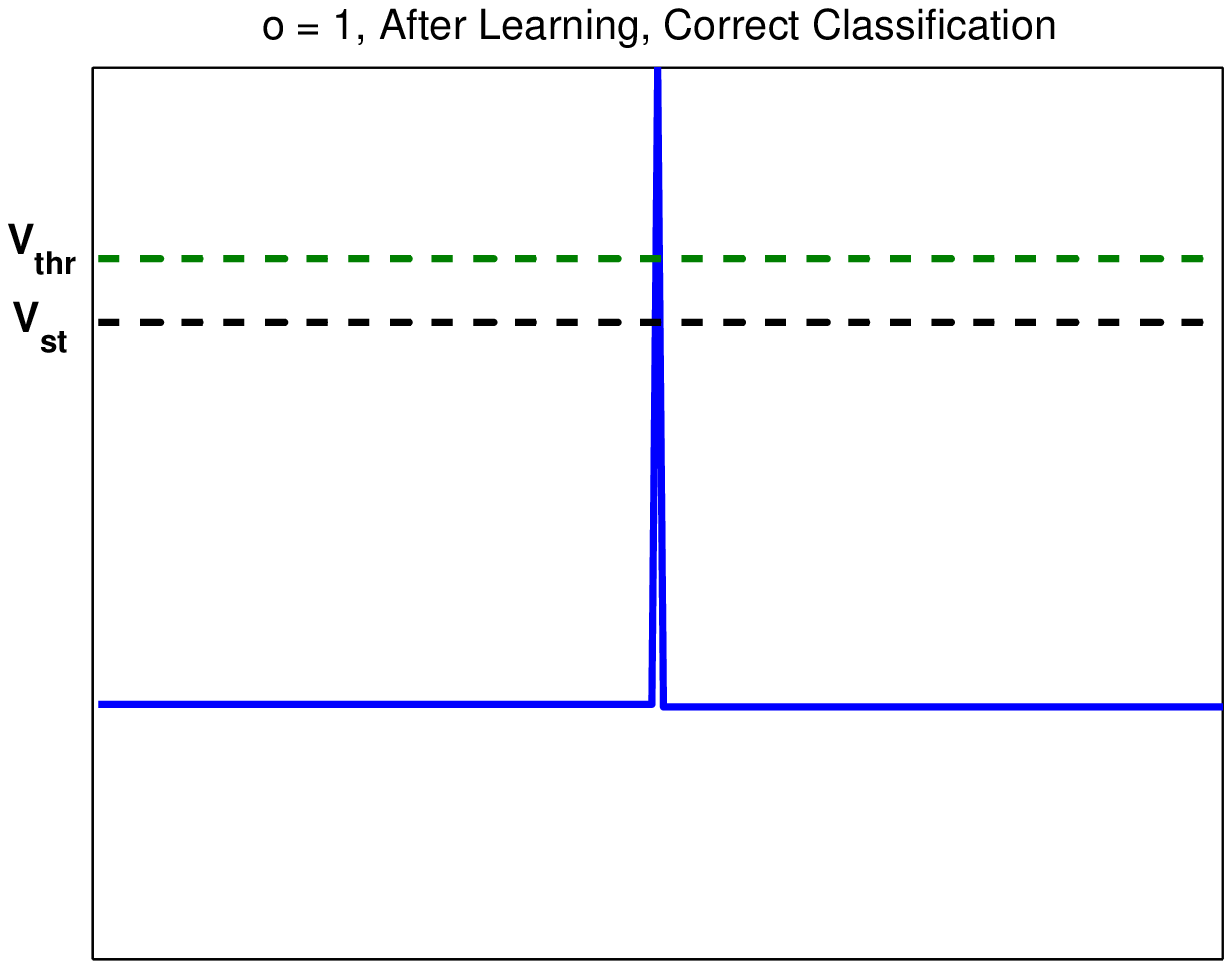}}
\\
\noindent 
\subfloat[]{\includegraphics[width=0.5\textwidth, height=0.25\textheight]{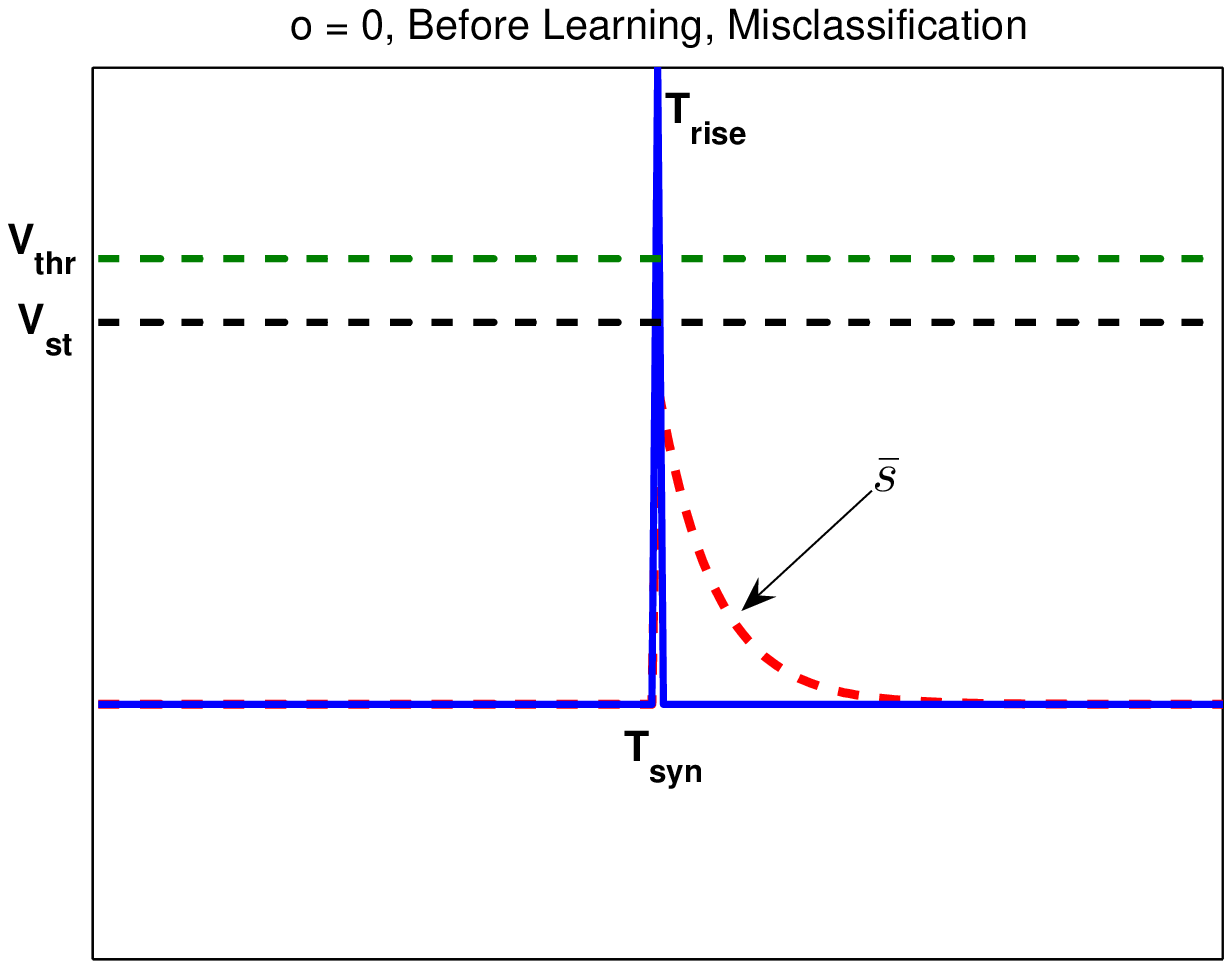}} 
\subfloat[]{\includegraphics[width=0.5\textwidth,height=0.25\textheight]{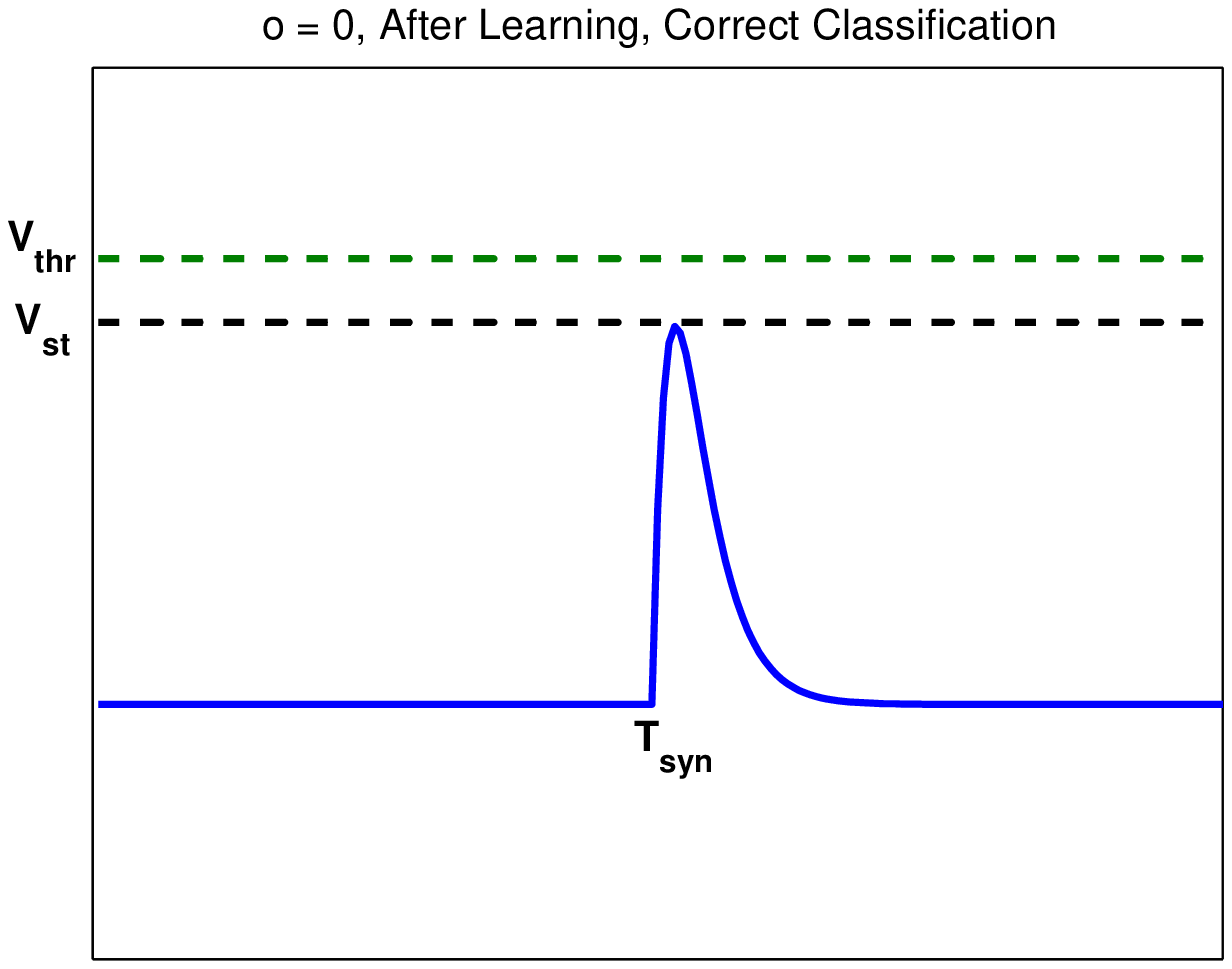}}
\end{center}
\caption[caption]{Spike-based learning for patterns belonging to the two classes: membrane voltage ($V_m(t)$) trace of the (+) neuron is shown for correctly classified patterns for $o=1$ (a) and $o=0$ (b) and for misclassified patterns before learning (c and e) and correct outputs after learning (d and f). } 
\label{fig:spike_learning_process}
\end{figure}

In order to understand how the $c_{ij}$ values computed for the BSTDSP rule approximate the $c_{ij}$ values of the RM learning (equations \eqref{eq:cell1} and \eqref{eq:cell2}), let us look at examples of all the cases -- when the model predicts the correct or wrong output ($y$) for $o=1$ or $o=0$ class of input pattern. Figure \ref{fig:spike_learning_process} shows the membrane voltage $V_m(t)$ of the (+) neuron under these different conditions. We consider cases where $x_{ij}=1$ since $\Delta c_{ij}=0$ for both neurons when $x_{ij}=0$. Further, we assume that there are $n$ active synapses out of $k$ synapses on the $j^{th}$ dendrite ($z_j=n)$ for each of the cases considered. Therefore, for $o=1$ and $y=1$, the correlation values can be written as:
\begin{align}
\Delta c^+_{ij}&=I_{b,out}^j(T_{syn}) \bar{r}(T_{syn}) - \gamma I_{b,out}^j(T_{rise}) \bar{s}_{ij}(T_{rise}) \notag \\
\Delta c^-_{ij}&=0
\end{align}
where the presynaptic spike arrives at time $T_{syn}$ and $V_m(t)$ crosses $V_{st}$ at time $T_{rise}$; $I_{b,out}^j(T_{syn})=b(z_j)=b(n)$. We assume that $\tau_{pre} \gg \tau_V$ which lets us write $I_{b,out}^j(T_{rise}) \approx I_{b,out}^j(T_{syn})=b(n)$. Therefore, $\Delta c_{ij}$ for this and other cases can be simplified as shown in Table \ref{table:cij_cases}. 

\begin{table}[h]
\renewcommand{\arraystretch}{1.3}
\caption{Comparison of $\Delta c_{ij}$ for the BSTDSP rule and RM.}
\begin{center}
\begin{tabular}{|L{2.5cm}|C{4cm}|C{4cm}|c|c|}\hline
   \multirow{2}{*}{Cases }    & \multicolumn{2}{|c|}{BSTDSP} & \multicolumn{2}{|c|}{RM} \\\cline{2-5}
   & $\Delta c^+_{ij}$  & $\Delta c^-_{ij}$  & $\Delta c^+_{ij}$ & $\Delta c^-_{ij}$  \\\hline
   $o=1, y=1$ (Figure \ref{fig:spike_learning_process}(a)) & $b(n)exp(-T_{syn}/\tau_{post}) - \gamma b(n)exp(-T_{rise}/\tau_{pre})$  &  $0$ &  $0$ &  $0$  \\\hline
   $o=0, y=0$ (Figure \ref{fig:spike_learning_process}(b)) & $0$  &  $b(n)exp(-T_{syn}/\tau_{post}) - \gamma b(n)exp(-T_{rise}/\tau_{pre})$ & $0$ &   $0$  \\\hline
   $o=1, y=0$ (Figure \ref{fig:spike_learning_process}(c)) & $b(n)exp(-T_{syn}/\tau_{post})$ & $- \gamma b(n)exp(-T_{rise}/\tau_{pre})$ & $b(n)$ & $-b(n)$ \\\hline
   $o=0, y=1$ (Figure \ref{fig:spike_learning_process}(e)) & $- \gamma b(n)exp(-T_{rise}/\tau_{pre})$ & $b(n)exp(-T_{syn}/\tau_{post})$ & $-b(n)$ & $b(n)$ \\\hline
\end{tabular}
\end{center}
\label{table:cij_cases}
\end{table}

We can equate the $\Delta c_{ij}$ values of the two models for the correctly classified cases to compute the value of $\gamma$ as given by:
\begin{equation}
\gamma = \frac{exp(-T_{syn}/\tau_{post})}{exp(-\bar{T}_{rise}/\tau_{pre})} \label{eq:gamma}
\end{equation}
Here, we can estimate $\bar{T}_{rise}$ by taking the average of all the times at which $V_m(t)$ crosses $V_{st}$ for different patterns. From the misclassified cases, we can write that:
\begin{equation}
[\Delta c_{ij}]_{BSTDSP} = exp(-T_{syn}/\tau_{post})[\Delta c_{ij}]_{RM} \label{eq:cij_models}
\end{equation}
Therefore, $c_{ij}$ values for spike-based learning are smaller than the $c_{ij}$ values calculated for the RM. Since connection changes are done on the basis of the ranking of $c_{ij}$, therefore equation \eqref{eq:cij_models} suggests that for the two models, the learning process comprising modifying connections should be the same for a given set of input patterns and for the same initial connections. This will remain true if the assumptions used to derive the above relationship are satisfied, i.e. $\tau_{pre} \gg \tau_V$ and that the estimate of $T_{rise}$ is close to the actual value of $T_{rise}$ for each pattern. 

Figures \ref{fig:spike_learning_process}(d) and \ref{fig:spike_learning_process}(f) show the correct classification outputs after learning where teacher signals are not available. Hence for $o=1$, in the absence of initial hyperpolarization, $V_m(t)$ starts from $0$ and can easily cross $V_{thr}$ thereby providing robustness against noise. During learning, $V_m(t)$ is encouraged to cross a lower threshold $V_{st}$, which ensures a margin for learning class (+) patterns, given by:
\begin{equation}
\delta^{+}_{spike} = (V_{st} - V_{reset}') - V_{thr}
\end{equation}
where $V_{reset}'$ is the voltage at the time $T_{syn}$ when an initial postsynaptic spike is present. For class (-) patterns, $V_m(t)$ does not cross $V_{thr}$ after learning while during learning $V_{m}(t)$ is reduced below $V_{st}$ providing a desired margin of:
\begin{equation}
\delta^{-}_{spike} = (V_{thr} - V_{st}) 
\end{equation}

For an unbiased classifier, the margins for the two classes are equal. Using this condition, the values of $V_{st}$ and $V_{reset}$ can be found. The value of the desired margin $\delta_{spike} = \delta^{+/-}_{spike}$ is determined by using the spike equivalent value of margin, $\delta$ (Section \ref{sec:modified}) after learning. This is done by computing the difference in the membrane voltage of the (+) and (-) neurons (analogous to $a^+-a^-$) when a single synapse is activated. This value multiplied by $\delta$ gives $\delta_{spike}$ since in the binary input case the synaptic strength is normalized to $1$. Since $\delta$ is a function of $m$ and $k$, we have also used different values of $\delta_{spike}$, which entails changing the values of parameters $V_{st}$ and $V_{reset}$ for different neuron configurations. 

\section{Results}
\label{sec:results}

Firstly, we discuss the procedure for training the RM and RMWM, followed by the method used to test the performance of these models.
\begin{itemize}
\item[(1)] Training on binary vectors: To demonstrate the learning process of the model, we used random patterns generated in a manner similar to \citet{mel_dendrite1}. The input patterns were chosen from a $d_o$ ($=40$) dimensional spherical Gaussian distribution. These patterns were randomly assigned to the positive ($o=1$) and negative ($o=0$) classes. The $d_o$-dimensional input patterns were mapped into a high $d$ ($=400$) dimensional space to incorporate sparse neural coding by using $n_{rf}=10$ non-overlapping binary-valued receptive fields (RFs) along each dimension. Hence, we generated binary vectors $\boldsymbol x \in \left\{0,1\right\}^d$ such that the probability of any one RF neuron being active is $0.1$, i.e. $P(x_i=1)=p=0.1$ $\forall i$. The binary patterns thus generated were used to train either the RM (Section \ref{sec:red_model}) or the RMWM (Section \ref{sec:modified}). The connection matrix learned after training process is used for testing the model. This method has the advantage of shorter simulation times as compared to the learning on spike inputs. It can be used in practice as well (if the desired input patterns are known) where the learned connection matrix can be downloaded to a neuromorphic chip.

\item[(2)] Testing on noisy spike inputs: We have tested the noise sensitivity of the RM and RMWM on noisy spiking versions of the binary input vectors used to train the models. Henceforth testing the model will refer to testing the noise tolerance of the model except for the classification of benchmark datasets, where the generalization performance is evaluated. We introduce noise in the inputs by using Poisson spike trains or jittered single spikes. We used Poisson spike trains because we wanted to model a realistic case where our classifier receives inputs from a noisy spiking sensor or from biological neurons in the case of brain-machine interfaces. The two types of spiking inputs used for testing are generated as: 1) for rate encoded spike patterns, binary values of $1$ and $0$ are mapped to Poisson spike trains with mean firing rates of $f_{high}$ and $f_{low}$ respectively, similar to \citet{mitra_spikeclassify} and 2) single spike patterns are generated by mapping binary input `$x_i=1$' to a single spike arriving randomly within $[T_{syn}-\Delta/2,T_{syn}+\Delta/2]$ and `$x_i=0$' to no spike, where $T_{syn}$ is a fixed time within the stimulus duration $T$. The testing on spike patterns utilizes the spike-based model (Section \ref{sec:arch}) with the learned connection matrix.
\end{itemize}

The results obtained are organized as follows:
\begin{itemize}
\item[(1)] Firstly, we discuss the classification performance of RM. These results were presented earlier in \citet{ijcnn_dendrite}, where we showed that the performance of RM on noisy spike trains is much worse than the training performance on binary vectors. Hence, RM suffers from poor noise tolerance. 

\item[(2)] Secondly, we show that the margin-based RMWM leads to improvements in both the training performance and the noise sensitivity. 

\item[(3)] Next, the classification results from the BSTDSP rule are presented, where we also compare the performance of the abstract RMWM with that of our bio-realistic BSTDSP model.

\item[(4)] Finally, the performance of RMWM is compared against SVM and ELM methods on $3$ different databases.

\end{itemize}

Simulation results are presented for $P=500$, $700$ and $1000$ input patterns using $m=10$, $20$ and $50$ dendrites and $k=5$, $10$, $15$, $25$ and $50$ synapses on each dendrite, unless stated otherwise. The learning parameters are $n_R=n_T=25$. The parameters for testing on spike patterns using the spike-based model and the parameters for the BSTDSP model are summarized in Tables \ref{table:spike_params} and \ref{table:bstdsp_params} respectively. These parameter values were chosen in accordance with the validity of the RM as discussed in Section \ref{sec:validity}. All the results are obtained by averaging across $5$ simulation trials.

\begin{table}[h]
\renewcommand{\arraystretch}{1.1}
\caption{Parameters for Spike-Based Model}
\begin{center}
\begin{tabular}{|c|c|c|c|c|c|c|c|c|}\hline
   $f_{high}$    &  $f_{low}$  & $\tau_r$  & $\tau_f$  & $I_0$  & $V_{thr}$  & $R$  & $C$  & $T$ \\\hline
   $250$ Hz      &   $1$ Hz    &  $2$ ms   &   $8$ ms  &   $2.12$  &  $10$ mV  &   $10$ M$\Omega$  & $5$ nF  & $200$ ms \\\hline
   
\end{tabular}
\end{center}
\label{table:spike_params}
\end{table}

\begin{table}[h]
\renewcommand{\arraystretch}{1.1}
\caption{Parameters for BSTDSP Model}
\begin{center}
\begin{tabular}{|c|c|c|c|c|}\hline
  $\tau_V$   &  $\tau_u$ &  $\tau_{pre}$  &  $\tau_{post}$  & $T_{syn}$\\\hline
  $4$ ms     & $80$ ms   &  $10$ ms       &  $50$ ms        & $100$ ms \\\hline
   
\end{tabular}
\end{center}
\label{table:bstdsp_params}
\end{table}

\subsection{Performance of RM}
\label{sec:rm_results}
First, we present the results of training the RM with binary vectors and then testing its robustness to noisy spike inputs.

\subsubsection{Training Performance}
\label{sec:train_rm}
\begin{itemize}
\item[(I)] \textbf{Comparison of L- and NL-neurons}

In the first experiment, we compare the performance of L- and NL-neurons in a binary classification task. Both the classifiers were trained on $P=500$ and $1000$ input patterns and their performances were compared. As shown in Figure \ref{fig:MAE_iter}, the MAEs decrease for both neurons as learning progresses. However, the NL-neuron achieves a lower error than its linear counterpart and this difference in error grows as the classifier is trained on larger number of patterns as seen from the performance on $1000$ versus $500$ training patterns.

\begin{figure}[!t]
\centerline{
\includegraphics[width=0.6\textwidth]{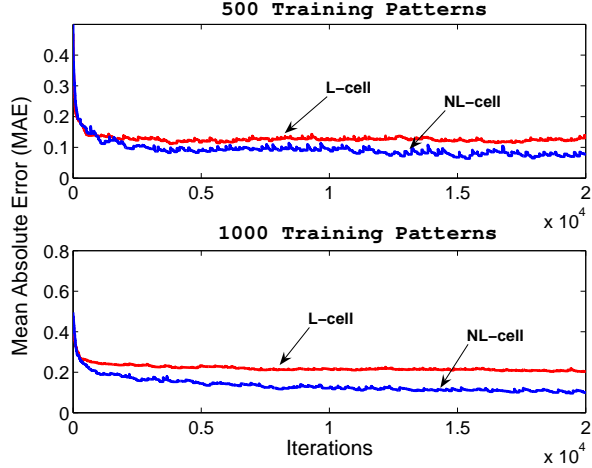}}
\caption{ Classification error as measured by the MAE of the L-neuron and NL-neuron. The plot shows the MAE as function of learning iteration for 500 and 1000 training patterns, number of dendrites, $m=20$.}
\label{fig:MAE_iter}
\end{figure}

In order to understand the effect of learning on the connection matrix, we investigated whether the input connections learned by a neuron are able to capture the correlations present in the input patterns. If $\boldsymbol x^{+}_{m}$ and $\boldsymbol x^{-}_{m}$ denote the $m^{th}$ binary vector belonging to class (+) and (-) respectively, then $[(\boldsymbol x^{+}_{m}).(\boldsymbol x^{+}_{m})^T]$ computes the second-order correlations present in the (+) pattern. $R=[\sum_{m=1}^{P^+}(\boldsymbol x^{+}_{m}).(\boldsymbol x^{+}_{m})^T]/P^+ - [\sum_{m=1}^{P^-}(\boldsymbol x^{-}_{m}).(\boldsymbol x^{-}_{m})^T]/P^-$ is a $d\times d$ matrix for $d$-dimensional patterns with $d^2$ entries of which the $d(d+1)/2$ values in the upper triangle and the diagonal are unique. $P^+$ and $P^-$ denote the total number of patterns in class (+) and (-) respectively. The entries in $R$ can be ordered according to decreasing magnitude which provides a unique ranking $O=\left\{ (r_1,c_1), (r_2,c_2), \ldots \right\}$ of the $d(d+1)/2$ entries in the upper triangle and the diagonal, where $r_i$ and $c_i$ denote row and column index respectively. Lower ranks correspond to high positive value entries in $R$, which denote $x_px_q$ combinations that are specific to pattern class (+) and not in pattern class (-). Similarly, high ranks correspond to high negative value entries in $R$ for those $x_px_q$ combinations that uniquely define pattern class (-). Similarly, $[(\boldsymbol W^{+}_{j}).(\boldsymbol W^{+}_{j})^T]$ gives the correlation matrix for the weights on a dendrite of the (+) neuron, where $\boldsymbol W^{+}_{j}$ is the $d$-dimensional weight vector of the $j^{th}$ dendrite, $\boldsymbol W^{+}_{j} \in \mathbb Z_{+}^d$, where $Z_{+}^d$ is a $d$-dimensional set of non-negative integers.

\begin{figure}[!t]
\centerline{
\includegraphics[width=1\textwidth]{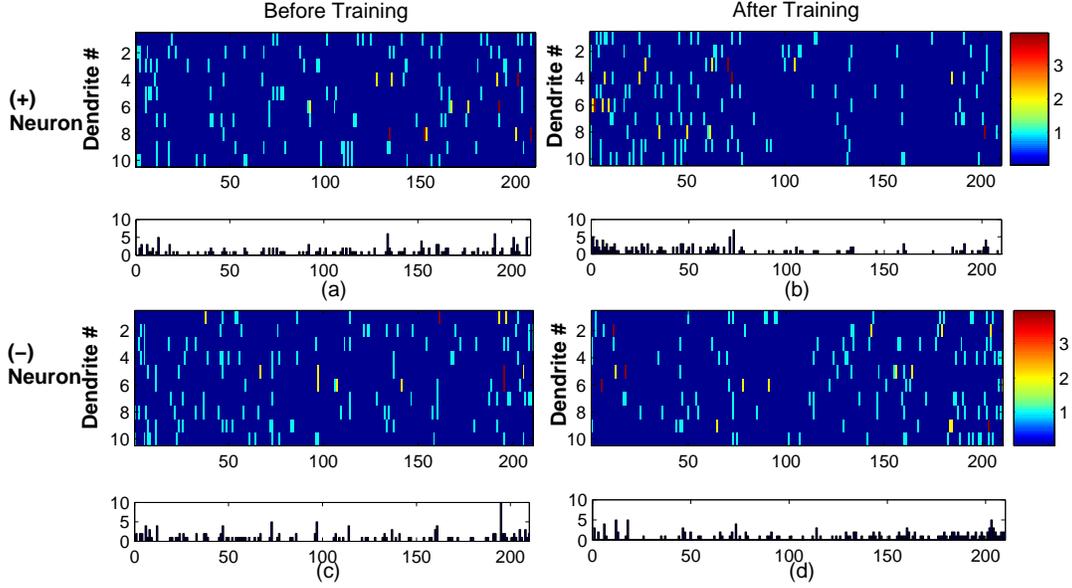}}
\caption{ $R_{W}$ values of the weight correlation matrix (x-axis) for (+) and (-) neurons with $m=10$ dendrites (y-axis) before and after learning. The histograms plotted by summing the correlation values for each input pair for all the dendrites are shown at the bottom of each $R_{W}$ matrix. }
\label{fig:W_corrs}
\end{figure}

Figure \ref{fig:W_corrs} shows how the input correlations learned by the (+) and (-) neurons match with those present in the input patterns. In this example, neurons with $m=10$ dendrites and $k=5$ synapses per dendrite were trained to classify $P=100$ binary patterns where $d=20$. $R_{Wj}$ is obtained by ordering the entries in the correlation matrix $[(\boldsymbol W_{j}).(\boldsymbol W_{j})^T]$ according to $O$ and is plotted for each dendrite of the (+) and (-) neurons before and after training. Each pixel shows the correlation value between a particular pair of input connections formed on a dendrite. For $d=20$, the total number of unique $x_px_q$ combinations is equal to $210$, and hence x-axis ranges from $1-210$ and y-axis ranges from $1-10$ for $10$ dendrites in each neuron. We have also shown the histograms obtained by plotting the sum of correlations for each connection pair for all the dendrites below each correlation matrix. Figures \ref{fig:W_corrs}(a) and \ref{fig:W_corrs}(c) show that the initial connections formed on the (+) and (-) neurons pick up the input correlations randomly, as evident from the correlation values evenly distributed along the x-axis. As seen in Figure \ref{fig:W_corrs}(b), the (+) neuron has a much higher clustering of connections corresponding to the lower ranked $x_px_q$ combinations unique to (+) patterns while it learns very few of the correlations $x_px_q$ with high negative values (higher ranked) unique to the (-) patterns. Similarly, the clustered appearance toward the right end of x-axis in Figure \ref{fig:W_corrs}(d) indicates that the (-) neuron learns to capture more of the correlations present in the (-) patterns and fewer of those in (+) patterns. Hence, the proposed learning algorithm enables the connection matrix to pick out unique correlations in the data. For the binary classifier consisting of a pair of neurons, we have shown that each neuron prefers patterns belonging to a particular class by mimicking the input correlations present in that class.

\begin{figure}[!t]
\centerline{
\includegraphics[height=70mm,width=0.7\textwidth]{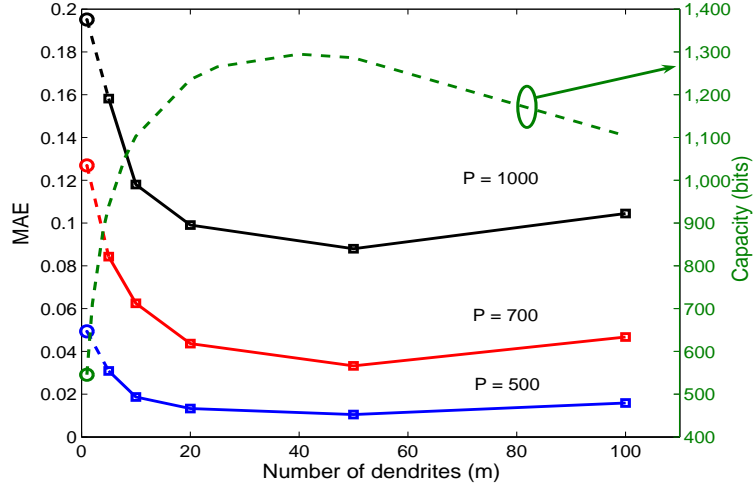}}
\caption{ MAE as a function of the number of dendrites, $m$. The L-neuron (circles) has higher classification errors than its nonlinear counterpart (squares). The optimal choice of $m$ for total number of synapses, $s=200$ is $m=50$. The dashed curve shows the theoretical capacity of the neuron for the same values of $m$ and $s$.}
\label{fig:fn_of_m}
\end{figure}

\item[(II)] \textbf{Effect of number of dendrites}

Next, we tested the performance of the model as the number of dendrites, $m$ is varied. The total number of synaptic resources was kept constant at $s=200$. Figure \ref{fig:fn_of_m} shows the MAE versus $m$. The circles represent the errors for the L-neuron with $m=1$ and the NL-neuron errors are denoted by squares ($m>1$). The performance of both types of neurons decreases with larger number of input patterns. However, the L-neuron performs much worse than the NL-neuron with classification errors of $19$\% against $9$\% for $1000$ patterns. Further the NL-neuron errors decrease as the number of branches increases, which can be explained by the fact that as $m$ increases, a larger number of dendritic functions can be selected from the $f$ possible functions obtained by drawing $k$ synapses from $d$ input lines (Section \ref{sec:theo_cap}). These nonlinear branch functions can well approximate the nonlinear classification boundary of the problem as explained in the example in Section \ref{sec:red_model}. As the number of dendrites continues to increase, the number of branch functions that can be drawn from $f$ functions reduces if $m$ increases beyond a certain value. As seen in Figure \ref{fig:fn_of_m}, the optimal value of $m$ is $50$ for $s=200$. The theoretical capacity of the neuron is also plotted on the right-hand y-axis (dashed), calculated by taking the logarithm of the combinatorial expressions given in Section \ref{sec:theo_cap}. As the capacity increases, more patterns can be learned and can be correctly classified. Hence, classification errors should decrease, which is in accordance with the empirical results. Similar to the measured classification errors, the theoretical capacity also peaks around $m=50$ and further reduces as $m$ increases. Therefore, as the number of dendrites grow, the classification performance of a NL-neuron with fixed synaptic resources ($s$) first increases to a peak value and then falls with further addition of dendrites.  

\item[(III)] \textbf{Effect of number of synapses per dendrite}

\begin{figure}[!t]
\centerline{
\includegraphics[height=70mm,width=0.7\textwidth]{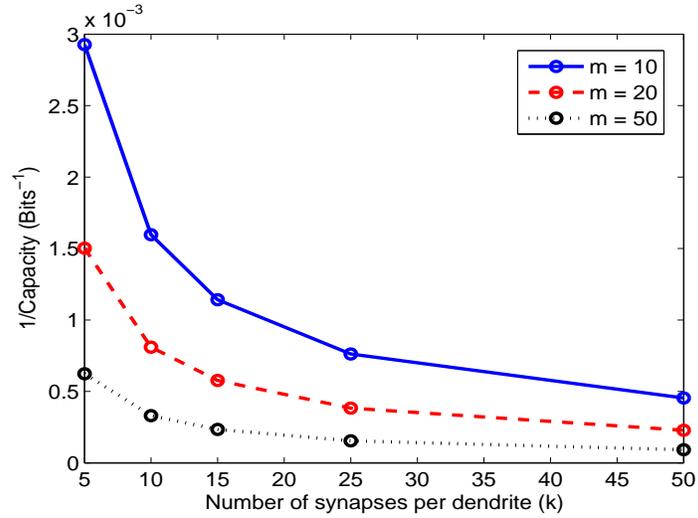}}
\caption{ Reciprocal of capacity of a neuron as a function of number of synapses per branch ($k$) for varying number of branches ($m$). Capacity increases as the total number of synapses $m\times k$ increases.}
\label{fig:Cap_mplots}
\end{figure}

We also determined the performance of NL-neuron as a function of number of synapses per branch ($k$) since it is a free parameter for an AER-based hardware system that can be easily adjusted. The reciprocal of theoretical capacity of a neuron as a function of $k$ is plotted in Figure \ref{fig:Cap_mplots}. It shows that for a neuron with a fixed number of dendrites, the inverse capacity decreases and therefore capacity increases as the number of synapses per dendrite is increased and as more dendrites are added, the neuron can learn more patterns. Figure \ref{fig:m_plots} shows classification errors measured empirically. The errors decrease as $k$ increases for a fixed value of $m$, which is due to the presence of higher number of total synapses $s=m\times k$. The larger number of patterns are more difficult to train and hence lead to higher errors. As larger number of dendrites are used, the errors decrease further and for a fixed number of synapses per branch, for example $k=25$, the classification error for $1000$ patterns reduces from $11.2\%$ for $m=10$ by a factor of $2$ for $m=20$ and of $6.5$ for $m=50$. Therefore, the empirical results agree with the theoretical capacity predictions that is, capacity increases with higher number of synapses, more patterns can be correctly classified. 

\begin{figure*}[t]
\centerline{
\includegraphics[width=1\textwidth]{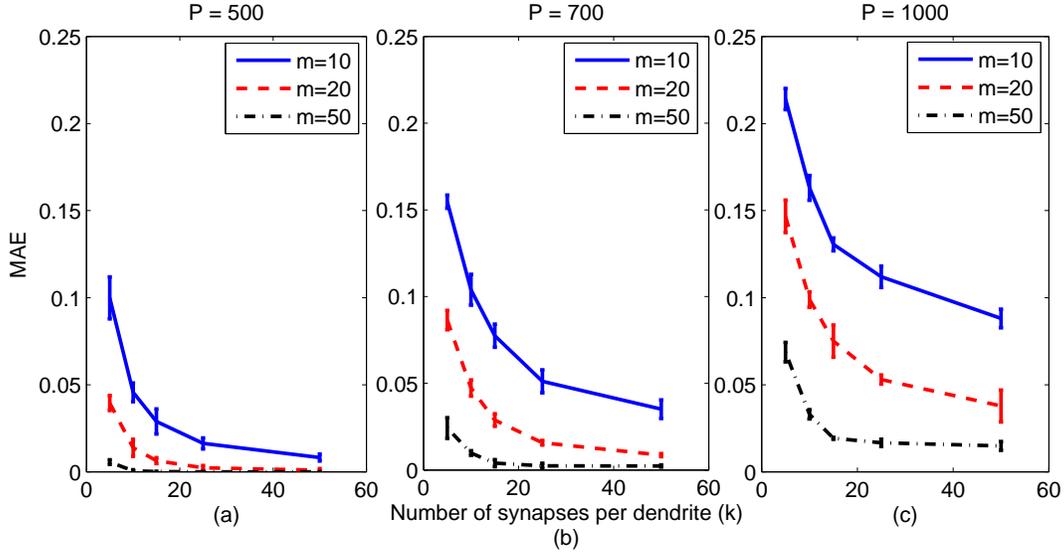}}
\caption{MAE as a function of $k$, the number of synapses per branch in a NL-neuron for binary inputs. $500$, $700$ and $1000$ patterns were trained on neurons with $m=10$, $20$ and $50$ branches. The errors decrease as the number of synapses increases.}
\label{fig:m_plots}
\end{figure*}

\end{itemize}

\subsubsection{Noise Sensitivity Performance}
\label{sec:test_rm}

\begin{figure}[!t]
\centerline{
\includegraphics[width=0.6\textwidth]{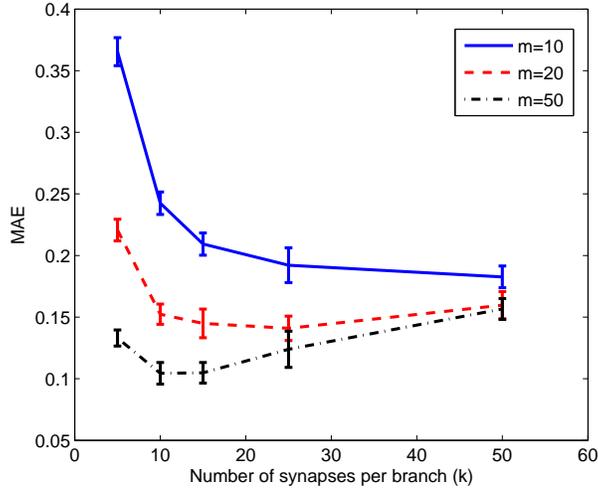}}
\caption{Testing performance for spike patterns as a function of $k$. The errors are much higher than that for the corresponding binary input vectors (Figure \ref{fig:m_plots}(c)).}
\label{fig:spike_errors}
\end{figure}

The spike-based model with the same connection matrix was used to classify spike patterns of mean firing rates. The classification performance of RM on spike patterns was measured as a function of $k$. Figure \ref{fig:spike_errors} shows that the classification errors for spiking inputs are much higher than that for the mean rate inputs on which the model was trained. It can be seen that the testing errors on noisy spike train inputs are $2-5$ times larger than that of the training errors for binary input vectors (Figure \ref{fig:m_plots}(c)). In contrast to the training errors decreasing with $k$, we observe that the testing errors increase as $k$ is increased.

The increased errors can be attributed to the fact that as number of branches ($m$) increases, the linear summation of dendritic branch outputs leads to a very high average value of the input currents $I_{in}^+$ and $I_{in}^-$ ($I_{in}=\sum_{j=1}^{m} I_{b,out}^{j}=\sum_{j=1}^{m} I_{b,in}^2/I_{thr}$) to the (+) and (-) neurons respectively. This causes the small differences between $I_{in}^+$ and $I_{in}^-$ to become indistinguishable for the (+) and (-) neurons due to the finite precision of the spike-based model used for testing as compared with the infinite precision of the abstract RM used for training. We also see from Figure \ref{fig:spike_errors} that the errors increase as the number of synapses per branch increases. This behaviour of the model can be explained by the fact that the use of higher dendritic threshold $I_{thr}$ to prevent branch saturation at larger $k$ reduces the $I_{in}^+$ and $I_{in}^-$ currents due to a spike pattern and therefore leads to decreased discriminability of the input currents to the spike-based (+) and (-) neurons. Next, we investigate how these issues are rectified using RMWM. 

\subsection{Performance of RMWM}
\label{sec:results_rmwm}
Here, we present the results of training the RMWM with binary vectors and then testing its noise sensitivity.

\subsubsection{Training Performance with $g_{margin}()$ function}
\label{sec:train_rmwm}

\begin{figure}[!t]
\centerline{
\includegraphics[width=0.6\textwidth]{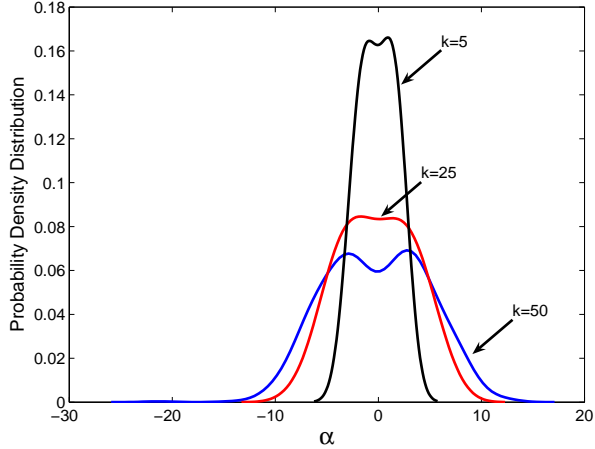}}
\caption{Determination of $\delta$ for $g_{margin}()$ function. The distribution of $\alpha$ for misclassified spiking inputs. $P=500$ and $m=20$.}
\label{fig:gdelta_dist}
\end{figure}

We present the results of applying the method discussed in Section \ref{sec:algo_modified} to determine the value of $\delta$ for the $g_{margin}()$ function. The model was trained on $P=500$ binary input patterns and then tested on the spike versions of these patterns. The $\alpha$ values for all the misclassified inputs were recorded. This experiment was repeated for NL-neurons with different number of branches and number of synapses per branch ranging from $k=5$ to $50$. The probability density distribution (pdf) of $\alpha$ as a function of number of synapses on each dendrite is shown in Figure \ref{fig:gdelta_dist} for $m=20$ and three values of $k$. We can see that the peak of the pdf of $\alpha$ varies with both $k$ and $m$ (Effect of `m' not shown in the Figure). Therefore, for a neuron with $20$ branches and around $5-50$ synapses per branch, $\delta_0$ can be set to around $20$ and then gradually reduced during training. Similar $\delta$ values can be used for training the NL-neurons with the same values of $m$ and $k$ on a classification dataset consisting of binary input patterns.

\begin{figure}[!t]
\centerline{
\includegraphics[width=0.6\textwidth]{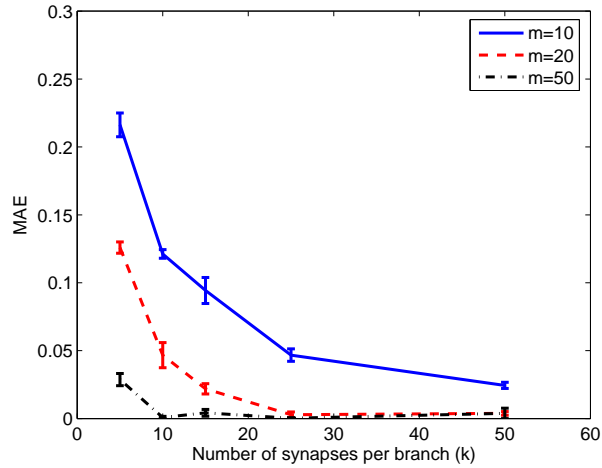}}
\caption{MAE as a function of k for the $g_{margin}()$ function. The training was done with $\delta_0$ set to $25$ and the input patterns were tested using $g()$ function. The performance improves for increasing $k$ and $m$.}
\label{fig:static_newg}
\end{figure}

Next we utilized the $g_{margin}()$ function (Figure \ref{fig:g_fns}(b)) to test if the performance of the classifier improves. The model was trained on $1000$ patterns for different values of $k$. The $g_{margin}()$ function was used during the training process with $\delta_0=25$. After the training was completed, the memorized patterns were recalled using $g()$ function. As shown in the Figure \ref{fig:static_newg}, the classification errors reduce for all neuron configurations. As compared to training errors of RM (Figure \ref{fig:m_plots}(c)), the errors corresponding to $g_{margin}()$ are smaller by factors of $2-5$ and for large $k$ values ($>25$), the errors reduce drastically by factors of more than $10-20$. Hence, the improved classification performance is observed because the model is trained to generate a decision boundary with a margin $\delta$. We will also present the results of improved noise tolerance achieved using the $g_{margin}()$ function.

\subsubsection{Improving Noise Sensitivity}
\label{sec:test_rmwm}

Here, we utilize the RMWM to improve the noise sensitivity by rectifying the problems discussed in Section \ref{sec:test_rm} in a step-wise manner. 

\begin{figure}[!t]
\centerline{
\includegraphics[width=0.7\textwidth]{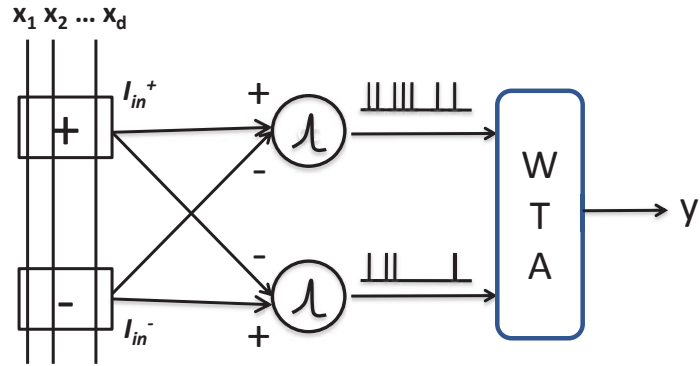}}
\caption{Architecture showing two neurons generating the output of the model in response to current differences ($I_{in}^+ - I_{in}^-$) and ($I_{in}^- - I_{in}^+$).}
\label{fig:2neurons}
\end{figure}

\begin{itemize}
\item[(a)] In order to balance the effect of high average values of the input currents for large values of $m$, we adopted the following strategy. The current differences ($I_{in}^+ - I_{in}^-$) and ($I_{in}^- - I_{in}^+$) are used as inputs to the two neurons (Figure \ref{fig:2neurons}). As shown in Figure \ref{fig:spike_issues_rect}(a), the reduction in classification error is most significant, by a factor of $1.2$ for $m=10$ as compared to when testing results were obtained without the differential input (Figure \ref{fig:spike_errors}). The errors reduce for smaller $k$ while for larger $k$, the performance still drops with increasing $k$. The dashed plots correspond to the training performance of the RMWM as shown in Figure \ref{fig:static_newg}. As described next, the RMWM proposes to achieve noise tolerance to spiking inputs which is close to this optimal training performance obtained by the model.

\begin{figure}[t]
\centerline{
\includegraphics[width=1\textwidth]{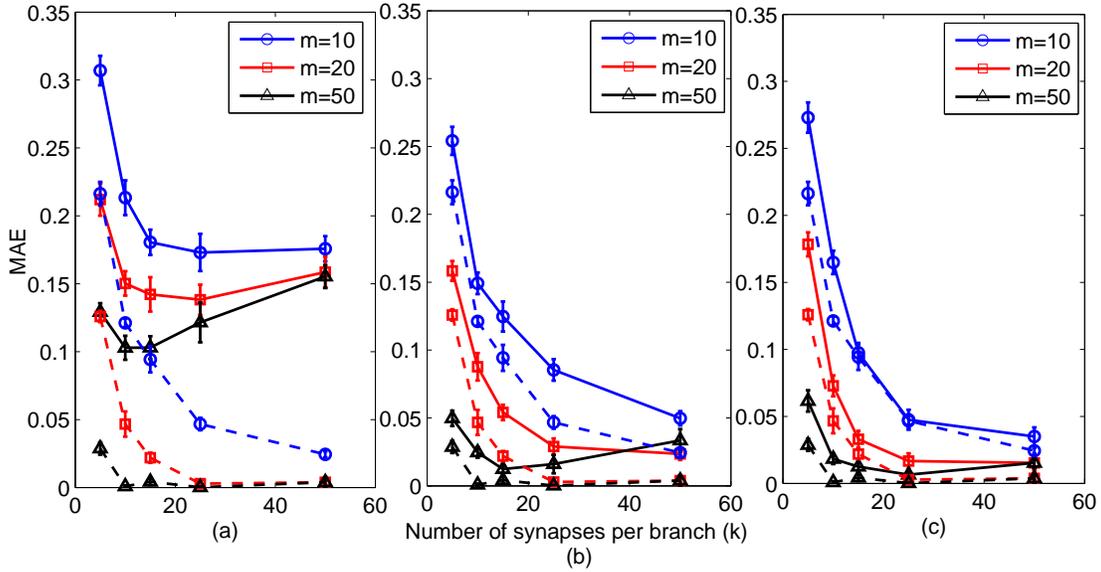}}
\caption{Performance on spike patterns improves as (a) differences in $I_{in}^+$ and $I_{in}^-$ used as input currents for the integrate and fire neurons to counter the high `m' effect; (b) $g_{margin}()$ used to improve the noise tolerance of the learning algorithm; and (c) inclusion of a dendritic nonlinear function $b_{leak}()$ improves the performance at larger $k$ values. $P=1000$, dashed plots correspond to the errors for binary vectors trained using the $g_{margin}()$ function.}
\label{fig:spike_issues_rect}
\end{figure}

\item[(b)] Next, we used the $g_{margin}()$ function to train the model on binary input vectors and then tested the performance on spiking inputs. This approach was used to improve the noise tolerance to the spike train versions. The errors computed using a NL-neuron are shown in Figure \ref{fig:spike_issues_rect}(b). As seen, the classification errors reduce for all combinations of $m$ and $k$ by factors of $1.5-6.5$ of that of errors obtained using RM and the performance levels match with the training performance (dashed plots) more closely than in Figure \ref{fig:spike_issues_rect}(a). These results suggest that training the model with $g_{margin}()$ function improves the robustness to noise by introducing a margin around the classification boundary. As seen in Figure \ref{fig:spike_issues_rect}(b), for $m=50$, the errors tend to rise with increasing $k$ as observed in the earlier results. Next, we discuss the approach used to solve this problem.

\item[(c)] We test whether the use of the dendritic nonlinearity, $b_{leak}()$ defined in equation \eqref{eq:bleak}, improves the performance for spike patterns at larger $k$ values. As shown in Figure \ref{fig:spike_issues_rect}(c), the errors at larger $k$ ($15-50$) are smaller by factors of $1.3-2.5$ as compared to errors obtained with the nonlinearity $b()$ (Figure \ref{fig:spike_issues_rect}(b)). However, there is no significant change in the performance for smaller $k$ values. This method further improves the test performance on noisy spiking inputs and the overall reduction in the classification errors for RMWM is by a factor of about $1.5-10$ of that of test errors for RM.

\end{itemize}

\begin{figure}[!t]
\centerline{
\includegraphics[width=0.6\textwidth]{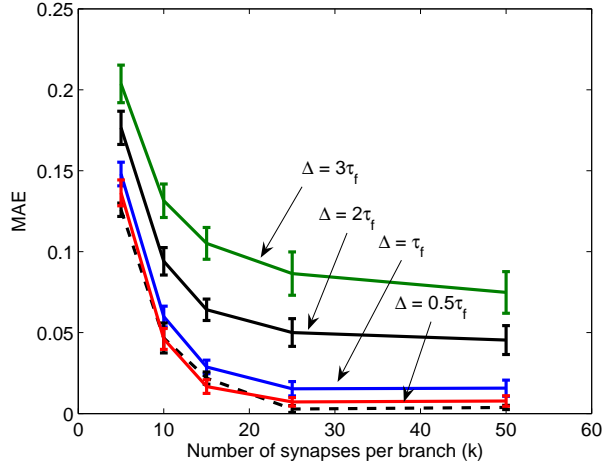}}
\caption{Testing performance for single spike patterns ($P=1000$) as a function of $k$ for $m=20$. The training (dashed) errors are about $2\%$ less than the testing errors for $\Delta\leq\tau_f$.}
\label{fig:syn_spike}
\end{figure}

\subsubsection{Performance on Single Spike Patterns}
\label{sec:test_single_spikes}

We also tested the performance of the model on patterns of single spikes. The input connections were learned using the RMWM on binary input vectors. The testing was done on $1000$ single spike patterns arriving within a time window of $\Delta=0.5\tau_f$, $\tau_f$, $2\tau_f$ and $3\tau_f$, where $\tau_f$ is the fall time constant of the postsynaptic waveform (equation \eqref{eq:kernel}) and $T_{syn}=100$ ms. The results shown in Figure \ref{fig:syn_spike} suggest that the model generalizes well on single spike patterns with testing errors about $1.2-4$ times that of the training errors (dashed) when $\Delta\leq\tau_f$, where $\tau_f=8$ ms. We found that this condition is satisfied for different values of $\tau_f=8$, $16$ and $24$ ms, the results for $\tau_f=16$ and $24$ ms are not shown in Figure \ref{fig:syn_spike}. As $\Delta$ increases, the testing errors increase because the spikes arriving at the synapses on the same dendrite are now more spread out in time leading to a reduced dendritic response as compared to the response learned with binary vectors.

\subsection{Performance of BSTDSP Model}
\label{sec:results_bstdsp}

\begin{figure}[!t]
\centerline{
\includegraphics[width=0.6\textwidth]{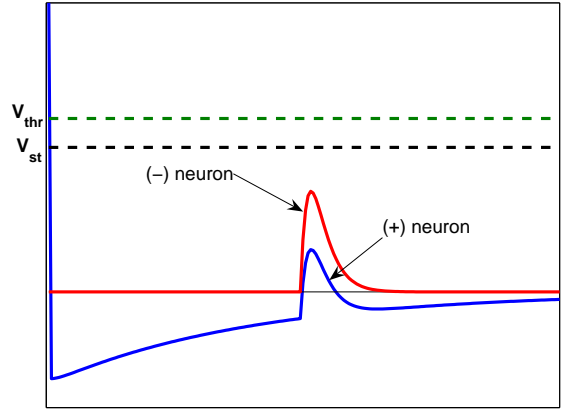}}
\caption{Membrane voltage $V_m(t)$ of (+) and (-) neurons when a class (+) pattern is presented. The pattern is not classified correctly since the (+) neuron fails to fire.}
\label{fig:demo_spike}
\end{figure}

We have evaluated the performance of our BSTDSP rule discussed in Section \ref{sec:bstdsp_model} on a two-class classification of spike patterns. The architecture used here is similar to the two neurons model shown in Figure \ref{fig:2neurons} where the input current into each neuron is the difference of currents with opposite signs. For training, we used random binary patterns of single spikes with $\Delta = 0$. In this case, we chose a low value of $V_{thr}$ ($=0.1$ mV) such that for the initial random connections, a neuron evokes a spike for about $65\%$ of the input patterns. The reason for choosing a low $V_{thr}$ value can be understood from Table \ref{table:cij_cases} where for the RM, any misclassification leads to $\Delta c_{ij}$ values (last two rows of Table \ref{table:cij_cases}) of opposite signs which contribute to the connection change in both (+) and (-) neurons after an epoch. This strategy of modifying both neurons limits the changes required in each neuron as compared to the case when only one neuron contributes at a time. Figure \ref{fig:demo_spike} shows a case where a (+) pattern is not classified correctly. However, only (+) neuron has non-zero $\Delta c_{ij}$ ($\Delta c^+_{ij}$ value for $o=1$ and $y=0$ in Table \ref{table:cij_cases}) while $\Delta c^-_{ij}=0$ because the (-) neuron does not fire a spike due to high $V_{thr}$. The use of a lower $V_{thr}$ will result in both $\Delta c^+_{ij}$ and $\Delta c^-_{ij}$ to be non-zero and of opposite signs.

\begin{figure}[!t]
\centerline{\includegraphics[width=0.5\textwidth]{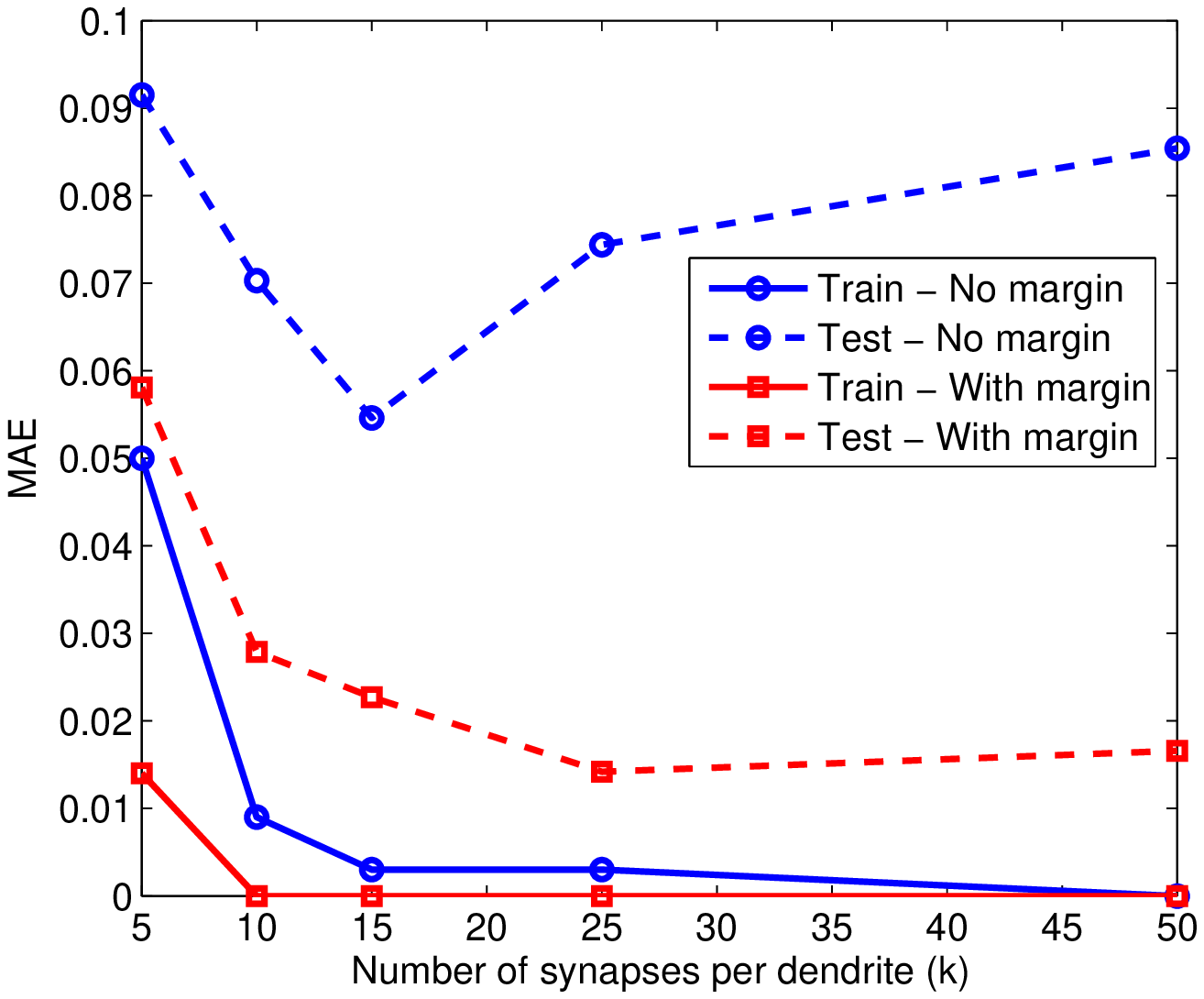}
\includegraphics[width=0.5\textwidth]{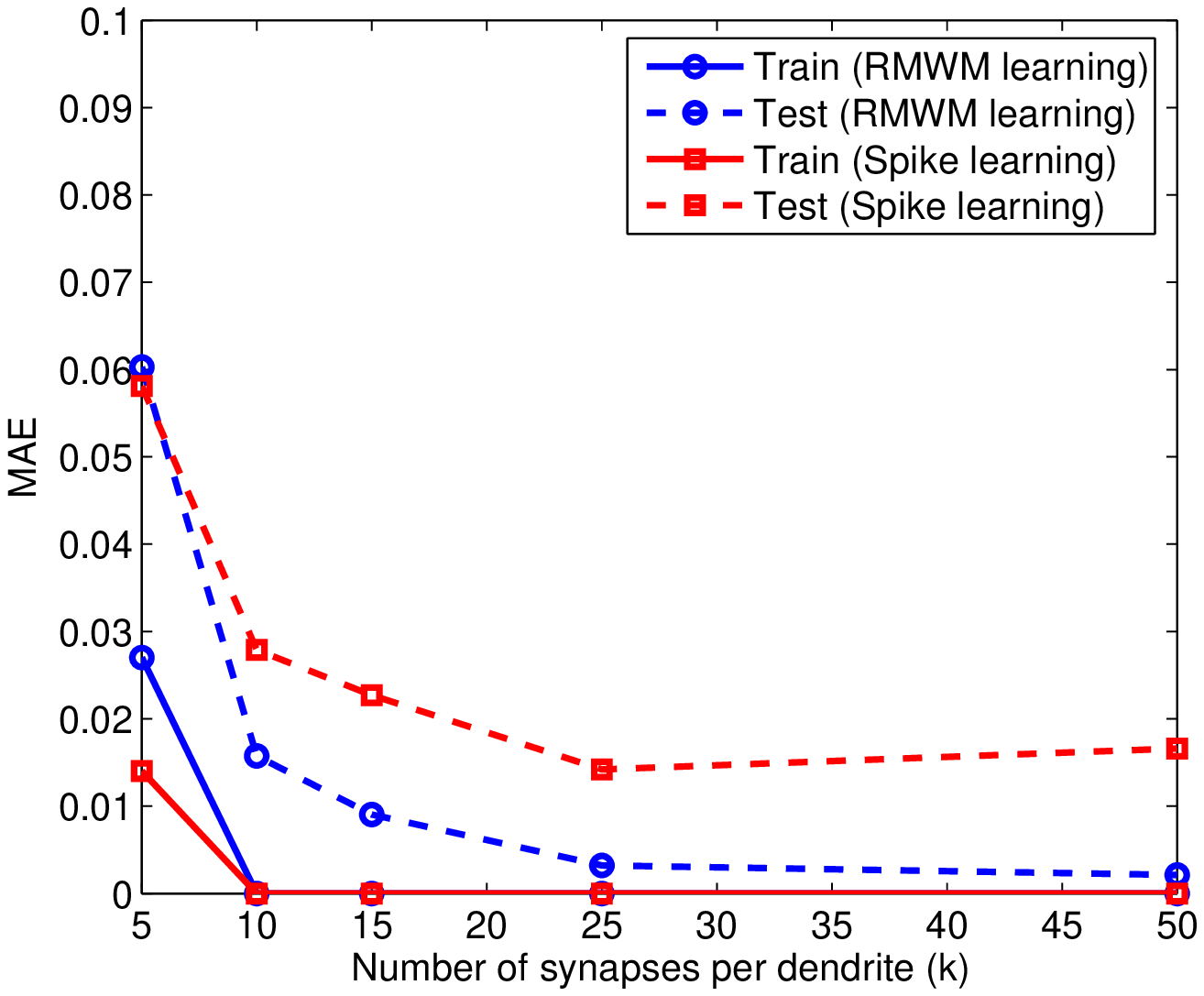}}
\caption{Classification performance of the BSTDSP algorithm. (a) Spike learning results when trained with (squares) and without (circles) margin. $P=500$ patterns of single spikes, $m=20$ (b) Comparison with RMWM learning (circles). The training errors are shown in solid and testing errors with dashed lines.}
\label{fig:spike_learning}
\end{figure}

The classification performance of the spike-based algorithm is shown in Figure \ref{fig:spike_learning}. The training errors are plotted as a function of $k$ for $P=500$ patterns. The model was also tested on noisy versions of the patterns of single spikes by adding a uniform random noise within the time window $\Delta = \tau_{f}$ around the spike time $T_{syn}$. Figure \ref{fig:spike_learning}(a) shows the performance when margin was not introduced, i.e. $\delta_{spike}=0$ (circles) and when margin was introduced by using $\delta_{spike}>0$ (squares), corresponding to different $k$ values. We can see that both training and noise testing performances improve significantly as a result of margin-based learning, similar to the results obtained for RMWM learning (Section \ref{sec:results_rmwm}). The errors for noisy spike inputs with margin classifier reduce by up to a factor of $5$ as compared to learning without margin. 

We also compared margin-based learning using spike inputs with the RMWM learning on mean rate inputs as shown in Figure \ref{fig:spike_learning}(b). The results show that the spike-timing based learning yields training errors (solid, squares) which are similar to that of training errors for RMWM (solid, circles). The test performance on noisy patterns of single spikes for the two instances of learning is shown with dashed plots. The test errors for spike learning are higher than the errors for RMWM learning by about $2.5$ times and this factor increases with $k$. These differences in the performance of BSTDSP from RMWM can be attributed to the differences in $\Delta c_{ij}$ calculations for the two schemes, which are not exactly same. As discussed in Section \ref{sec:learning}, the value of $\gamma$ as calculated in equation \eqref{eq:gamma} is based on the assumption that $\tau_{pre} \gg \tau_V$ which is not perfectly true in our case. This also results in the estimate of $T_{rise}$ used for calculating $\gamma$ to be different from the actual $T_{rise}$ for each pattern. Hence, the values of $\Delta c_{ij}$ are slightly different from the ideal values as shown in Table \ref{table:cij_cases}. Another discrepancy between the learning for BSTDSP and RMWM can arise from the fact that the margin setting is not exact. As discussed in Section \ref{sec:learning}, the margin for spike-based learning was set using $\delta_{spike} = \eta \delta$ where a wrong estimate of the constant $\eta$ will mean that $\delta_{spike}$ and $\delta$ are not truly analogous to each other, leading to different results. However, we can see that even with these differences in the learning for the two models, our proposed bio-realistic spike-based learning can achieve comparable errors with the more precise learning in the abstract model presented earlier. 

\subsection{Performance on Benchmark Datasets}
\label{sec:results_uci}

Finally, we tested the performance of the proposed algorithm on real-world benchmark classification tasks and compared the results with the performance of the SVM and ELM methods. The binary datasets include datasets on breast cancer (BC), heart disease (HEART) and ionosphere (ION). These datasets (Table \ref{table:data_specs}) are taken from the UCI Machine Learning Repository \citep{uci_data}. The performance of our improved RMWM with structural learning rule was evaluated for both binary vector patterns as well as spiking inputs, where each sample corresponding to a feature was mapped into ten non-overlapping density-matched RFs, which were generated from the probability distribution of all samples of that particular feature. The resulting binary samples were then converted into spike patterns by using the method discussed in Section \ref{sec:results}.

\begin{table}[h]
\renewcommand{\arraystretch}{1.3}
\caption{Specifications of Benchmark Binary Datasets.}
\begin{center}
\begin{tabular}{|l|c|c|c|}\hline
   Datasets    & \# features  & \# training samples  & \# testing samples\\\hline
   BC          &   $9$        &    $222$               &   $383$\\\hline
   HEART        &   $13$        &    $70$               &   $200$ \\\hline
   ION         &   $34$       &    $100$               &   $251$ \\\hline
\end{tabular}
\end{center}
\label{table:data_specs}
\end{table}

\begin{table}[h]
\renewcommand{\arraystretch}{1.3}
\caption{Performance Comparison of SVM, ELM and Dendrite Algorithms on Benchmark Datasets.}
\begin{center}
\begin{tabular}{|l|p{1.2cm}|c|p{0.8cm}|p{1.2cm}|c|p{0.8cm}|p{1.2cm}|c|p{1cm}|p{1cm}|}\hline
   \multirow{2}{*}{Datasets}    & \multicolumn{3}{|c|}{SVM} & \multicolumn{3}{|c|}{ELM} & \multicolumn{4}{|c|}{Dendrite Model}\\\cline{2-11}
   & \# neurons  & \# wts  & Acc. & \# neurons  & \# wts  & Acc. & \# dendrites  & \# wts  & Acc. (Binary) & Acc. (Spike)\\\hline
   BC          &   $24$        &    $240$        &   $96.61$ &   $66$        &    $660$       &   $96.35$ &   $20$     &    $204$     &   $96.01$ & $95.93$\\\hline
   HEART        &   $42$        &    $588$        &   $75.5$ &   $36$        &   $504$       &   $76.5$ &   $10$     &    $104$     &   $75.3$ &   $74.53$\\\hline
   ION         &   $43$         &    $1505$        &   $91.24$ &   $32$      &   $1120$      &  $89.64$ &  $50$    &    $404$   &   $89.22$ & $88.96$\\\hline
\end{tabular}
\end{center}
\label{table:perf_data}
\end{table}

The performance measures like number of neurons, number of weights and classification accuracy for different algorithms are given in Table \ref{table:perf_data} to compare the classification performance as a function of the computational complexity of the network used. The results for SVM and ELM are taken from \citet{suresh_ref}. The number of nonlinear dendrites of a neuron is analogous to the number of support vectors of the SVM and the number of hidden neurons for the ELM. The performance of different classifiers is also compared on the basis of synaptic resources used. The total number of weights in an $N$-input network (SVM and ELM) using $M$ neurons is given by $N\times M + M$. The total number of weights in the dendritic model with $m$ dendrites and $k$ synapses per dendrite for each neuron in the classifier is given by $s=m\times k$. The classification accuracy obtained in this work for spike inputs is slightly less ($0.1-0.8 \%$) than that for binary input vectors. The reduced accuracy is probably due to the noise in the Poisson spike trains and the effects of the integration and fire process of the neuron. As the results suggest the proposed model uses a similar number of dendritic processing units as the number of neurons used by SVM and ELM and achieves generalization performance within $1-2 \%$ of that of these algorithms. Moreover, our method achieves this performance by utilizing $10-50 \%$ fewer binary synapses, particularly for HEART and ION datasets, therefore rendering the model feasible for hardware implementation.

\section{Discussion}
\label{sec:discussion}

Here, we discuss the biological and machine learning perspectives of models which compute a higher-order function of the inputs, particularly polynomial functions. We also present neurobiological and computational evidences to suggest that these higher-order units can be mapped to active dendrites consisting of clustered coactive synapses and the fact that formation of such synaptic clusters is supported by the mechanism of structural plasticity. Finally, we compare the features of our model with that of other machine-learning classifier models based on similar concepts as ours.

\subsection{Sum-of-Products: Biological Perspective}
\label{sec:squareNL}

Our proposed model employs a static quadratic dendritic nonlinearity as approximation to the nonlinear dendritic processing due to voltage-gated ion channels. This simplification leads to an abstract neuron model in which the neuron output can be expressed as a sum of squaring subunit responses which can be further written as the sum of products of synaptic inputs on a dendritic subunit. The idea that the input-output behaviour of a complex dendritic tree can be reduced to a simple algebraic formula consisting of sum-of-squares is supported by an earlier compartmental model for disparity tuning in complex cells of primary visual cortex \citep{square_subunits}. This model consisting of simple-cell-like subunits mapped onto active dendrites, behaves like a hypothetical cell with squaring subunits. In another study, the firing rate of a complex biophysical pyramidal cell model was consistent with the mean rate predicted by an abstract two-layer neural network with sigmoidal subunits representing the thin terminal dendrites \citep{mel_2layerNw}. Hence, these and other studies \citep{mel_nips_learningdendrite, sigma_pi_comp} suggest that from a biophysical perspective, a pyramidal cell with varying density, spatial distribution and kinetics of voltage-gated ion channels can be well approximated by an abstract model of sum of nonlinear subunits. 

\begin{figure}[!t]
\centerline{
\includegraphics[width=0.7\textwidth]{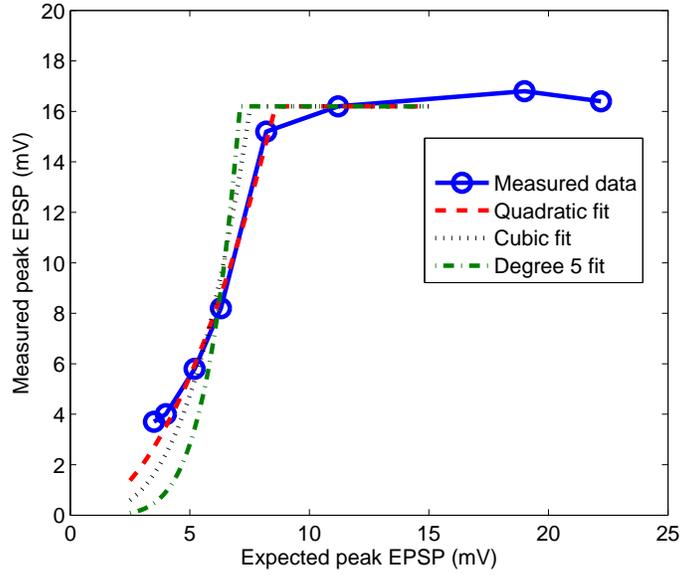}}
\caption{Expected vs actual peak voltage responses to stimulation of synaptic sites on the same dendrite (Reproduced from Figure 4 in \citet{mel_dendrite_b}) shown with circles. Rest of the curves show polynomial fits to the experimental data.}
\label{fig:b_fit}
\end{figure}

To investigate the particular form of nonlinear function which might represent the input-output behaviour of a dendrite, we look at the experimental study supporting the two-layer model of sum of subunits \citep{mel_dendrite_b}. In this study confocal imaging was combined with focal synaptic stimulation to examine the rules of subthreshold synaptic integration in thin dendrites of pyramidal neurons. Figure \ref{fig:b_fit} shows the measured and expected peak responses when two nearby sites on the same branch were stimulated simultaneously, where the expected response is the arithmetic sum of the responses to individual stimulations. The solid plot marked with circles corresponds to the measured responses. This curve has been reproduced from the data presented in Figure 4 in \citet{mel_dendrite_b} for single-pulse stimulation. Therefore, experimental evidence suggests that the dendritic response to synaptic inputs on that branch is governed by a saturating expansive nonlinearity. In order to determine the form of subunit nonlinearity, we have fit polynomials to the experimental data because it is easier to implement polynomial functions in a hardware system. We have used the polynomials $y=\beta x^2$, $y=\beta x^3$ and $y=\beta x^5$ with saturating effects, to fit the data, where the constant $\beta$ was chosen to get the best fit. As shown in Figure \ref{fig:b_fit}, the quadratic function provides the best fit as compared with degree $3$ and degree $5$ polynomials tested here. Hence, for the few data points obtained from biological measurements, we can consider the squaring function as a valid choice for the dendritic nonlinearity.

\begin{figure}[!t]
\centerline{
\includegraphics[width=0.5\textwidth]{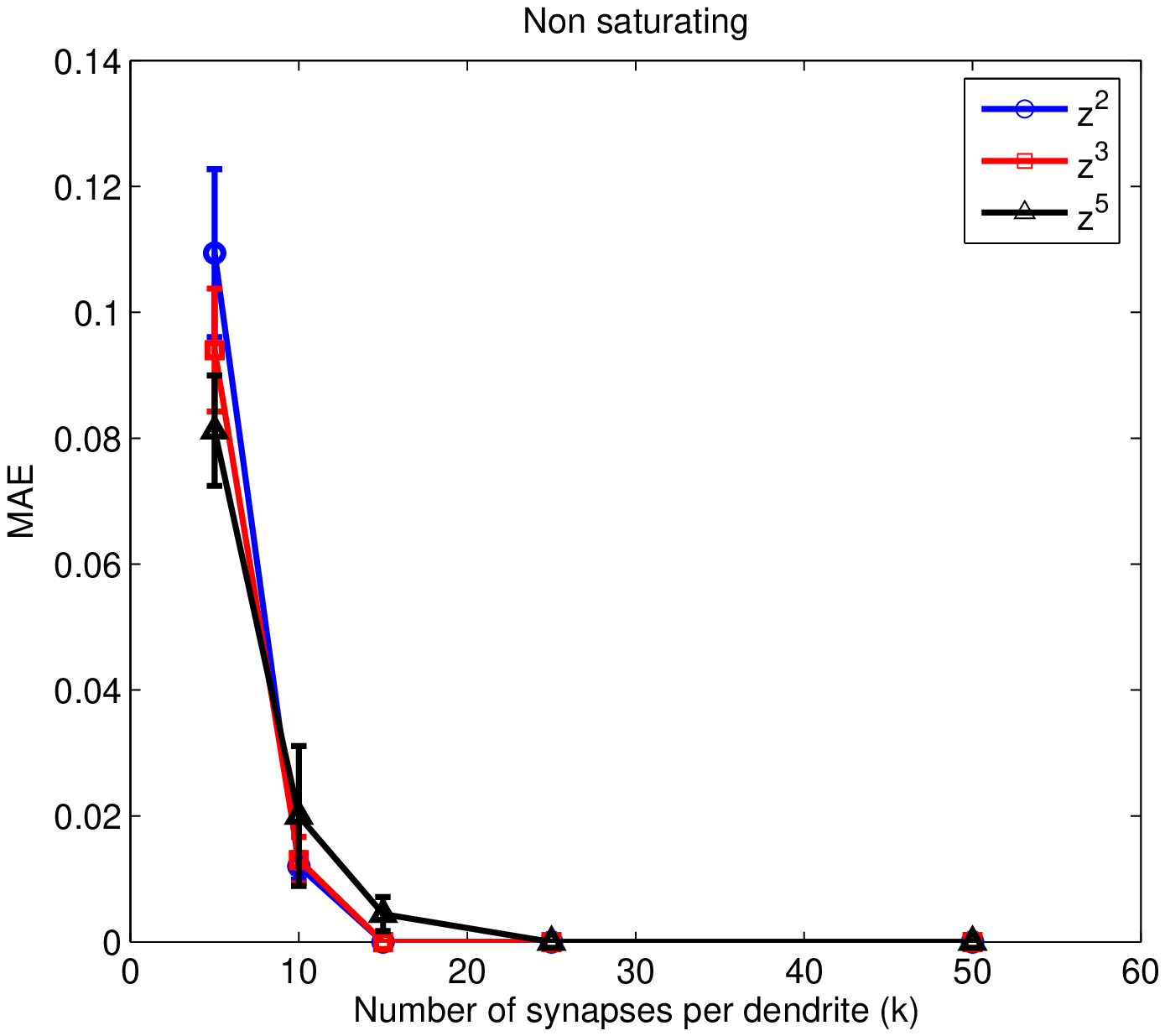}
\includegraphics[width=0.5\textwidth]{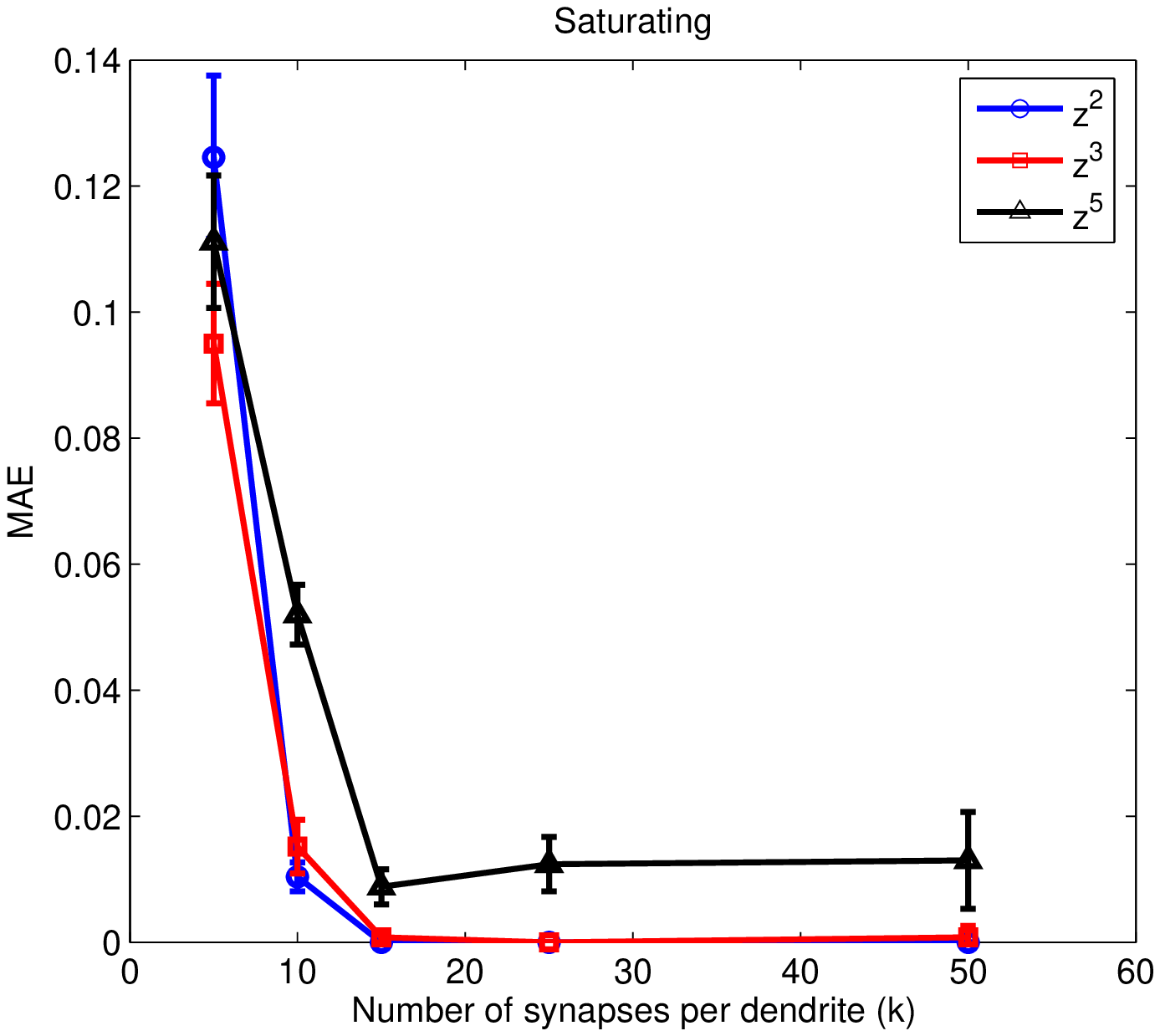}}
\caption{(a) Polynomial functions of different degrees used as branch nonlinearity $b()$. Saturation condition is not used here for learning $P=500$ patterns (b) Saturating polynomial nonlinearities used as $b()$ function.}
\label{fig:polyNLs}
\end{figure}

Further, we addressed this question from the perspective of performance in machine learning tasks. We tested the different dendritic polynomial nonlinearities used above in the random pattern binary classification task for our proposed model. As above, we compared the polynomials of degree $2$, $3$ and $5$ under the conditions when saturation is either enforced or not. The exact polynomial function used is given in the equation for $b_{leak}()$ (equation \eqref{eq:bleak}), where the exponent is $2$, $3$ or $5$. Moreover, the value of margin $\delta$ used in the $g_{margin}()$ function is determined for each form of nonlinearity used according to the method discussed in Section \ref{sec:algo_modified}. The results shown in Figure \ref{fig:polyNLs}(a) suggest that when there is no saturation on the nonlinearity, all three polynomials result in similar performance, that is improvement comes with more synapses on each dendrite. This indicates that the particular form of nonlinearity does not have any influence over the classification performance. This result can be explained by the fact that from the combinatorial expressions used to predict the storage capacity of the nonlinear neurons (Section \ref{sec:theo_cap}), we can see that the purpose of branch nonlinearity is to partition the multiple dendrites into independent computational units and therefore the type of nonlinearity used should not have an effect. Figure \ref{fig:polyNLs}(b) shows that when a fixed saturation level is applied, lower degree polynomials ($z^2$, $z^3$) yield better classification performance than a higher degree polynomial ($z^5$), which is due to more branches getting saturated faster in case of a higher power branch function. Finally, our choice of quadratic nonlinearity is supported by easier implementation in hardware, biological plausibility and better performance in a classification task.

\subsection{Sum-of-Products: Machine Learning Perspective}

It was shown in Section \ref{sec:red_model} that our proposed classifier uses a discriminant function of higher-order terms to perform nonlinear mapping for classification. Several higher-order networks have been proposed to solve nonlinear classification problems. These models utilize higher-order combinations of some of the input components to solve the problem. A method to construct higher-order networks involves incremental technique in which the number of free parameters is gradually increased \citep{pi_sigma}. A polynomial neural network utilizes this technique which is based on an evolutionary strategy of fitting a high-degree polynomial to the input components using a multilayered network \citep{pnn_gmdh}. This approach employs polynomial theory \citep{ivakh} to determine the polynomial description of optimum complexity which is necessary to achieve high prediction accuracies. Another method to generate higher-order functions of the input components involves a functional link network which estimates an unknown function as $\sum_{i}w_i \phi_i(\boldsymbol x)$, where $\boldsymbol x$ is the input vector and $\phi_i(\boldsymbol x)$ serve as basis functions such as products of inputs or sinusoidal functions. This method suffers from the need to use \emph{a priori} knowledge in order to choose optimal basis functions \citep{pi_sigma}.

A special case of this approach, when $\phi_i(\boldsymbol x)$ are products of input components, is discussed next. This category includes several models which employ higher-order correlations among input components to perform nonlinear mapping. Such higher-order networks can be trained using fast learning schemes like Hebbian learning rule or perceptron type learning. A sigma-pi network is a type of high-order network consisting of a hidden layer which computes the product of inputs, followed by a summing unit \citep{sigma_pi_book}. The output of a ``sigma-pi" unit is given by the following equation:
\begin{equation}
y=\sigma (w_0 + \sum_{j} w_j x_j + \sum_{j,k} w_{jk} x_jx_k + \sum_{j,k,l} w_{jkl} x_jx_kx_l + \cdots)
\end{equation}
where $\sigma()$ is a sigmoid-like activation function, $x_j$ is the \emph{j}th component of input vector $\boldsymbol x$, $w_{jkl}\ldots$ is a weight which captures the correlation between product of input components $x_j$, $x_k$, $x_l$, $\cdots$ and the output of the unit; and $w_0$ is an adjustable threshold. The drawback of higher-order networks consisting of such units is the combinatorial explosion of terms and hence weights needed to accommodate all possible $2^N$ correlation terms among $N$ inputs, where input $\boldsymbol x \in R^N$ . This exponential increase in the number of higher-order terms with the input dimension limits the applicability of the higher-order approach. Another higher-order network is the pi-sigma network (PSN) which computes the \emph{products of sums} of input components instead of \emph{sums of products} as in sigma-pi networks \citep{pi_sigma} and overcomes the problem of combinatorial increase in the number of weights. Durbin and Rumelhart \citep{prod_units} used product units which represent any generalized higher-order term and can learn which one to represent. These units can be incorporated in layered networks along with thresholded summing units.

\subsection{Correlation-based Synaptic Clustering: Biological Relevance and Computational Modeling}

Our hardware-friendly learning rule involves aggregation of correlated synapses on a dendrite, which is consistent with several experimental findings. These studies have demonstrated that clustering of synapses on dendritic branches might be a substrate for learning. This phenomenon suggests that single dendritic segments serve as functional subunits thereby expanding the neuronal properties significantly. This is in agreement with the predictions of the earlier computational studies. The mechanism of nonlinear dendritic integration leads to regenerative events including dendritic spikes, which have a stronger influence on neuronal firing than the activation of synapses located on different dendrites \citep{Larkum_review}. This evidence suggests clustering of coactive inputs within a dendritic subregion, which can lead to compartmentalization of various forms of signalling in dendrites, like electrical, biochemical and cellular \citep{Branco_review}. Based on these results, it has been proposed that at the single cell level, dendritic branches act as basic functional units for long-term memory storage \citep{dend_unit_protein}. One such study has reported that long-term potentiation (LTP) at a single spine can modulate the threshold for plasticity at neighbouring spines \citep{cross_ltp}. This `crosstalk' between plasticity at nearby synapses results in dendritic segments having similar properties, which could allow the memory traces to preferentially form at synapses clustered within these dendritic subunits \citep{dend_unit_protein}. \citet{plast_compartments} showed that synapses innervating the proximal and distal regions of basal dendrites are governed by different rules for the induction of LTP. This domain-specific plasticity mechanism may support different learning functions. 

The first demonstration that synaptic clustering might be a substrate for learning was given in the barn owl auditory system \citep{barn_owl_clust}. The authors reported that behavioural experience drives changes in the clustering of axo-dendritic contacts. A related study has shown that new spines emerge in clusters along dendrites as a result of the repetitive activation of the cortical circuitry during motor learning \citep{syn_clust_motor}. Synaptic clustering was also reported from a calcium imaging study which demonstrated that locally convergent inputs in spontaneously active networks lead to frequent synchronization of adjacent spines thereby facilitating dendritic compartmentalization \citep{synch_syn_inputs}. In addition to the structural clustering of synapses on dendrites, a novel form of plasticity at the dendritic level is provided by the excitability of the dendritic branches. A study by \citet{dend_coupling} showed that experience in enriched environments can lead to compartmentalized increase in dendro-somatic coupling. A recent study showed activity-dependent fine-tuning of synaptic inputs having temporally correlated activity in a developing neuron \citep{act_dep_clust}. This pattern of connectivity was predicted by a computational model \citep{mel_dendrite1} based on theoretical capacity measures. This neuron model consisted of multiple dendritic subunits, which acted as independent computational units. 

Several other computational models have hypothesized earlier that synapses on a dendrite are activated in clusters \citep{sigma_pi_book, SigmaPi_bio, mel_nips_learningdendrite, sigma_pi_comp}. In a sigma-pi model, the multiplicative interactions among neighboring synaptic inputs could underlie local dendritic computations such that these local dendritic regions act as product units whose outputs are then summed at the soma \citep{SigmaPi_bio}. In this study, it was argued that the output of a cluster of synaptic inputs on a dendrite can be AND-like, i.e. selective for co-activation of all the inputs. Mel and his group later presented several computational studies to elucidate the role of dendrites which can be incorporated in neuron models rendering them computationally more powerful \citep{mel_dendrite1}. A proposed abstract model neuron called the ``clusteron" maps low-order polynomial functions of inputs onto a dendritic tree receiving synaptic contacts from a set of afferent axons \citep{mel_nips_learningdendrite}. Later, it was hypothesized that a single neuron acts like a powerful multilayered network which performs nonlinear operations within spatially separated dendritic branches \citep{mel_infoPro}. In a more recent study, a two-layer dendrite model was proposed \citep{mel_2layerNw} in which different sets of inputs targeting functionally distinct dendrites are first processed nonlinearly and then the dendritic outputs are linearly integrated at the soma. 

\subsection{Structural Plasticity: Biological Relevance}

The learning process in our proposed neuron model with nonlinear dendrites involves formation and elimination of synapses to enable clustering of coactive synapses on a dendrite. \citet{SigmaPi_bio} proposed that a tangle of axons within a dendrite's receptive field facilitates the formation of clusters of synapses on local regions of dendritic arbor. The process of constructing the optimal connection matrix involves alterations to the cortical `wiring diagram', which requires structural plasticity at the axo-dendritic interface comprising synaptogenesis, axonal and dendritic growth and remodeling, and also retraction and reformation of dendritic spines \citep{rewiring_chklo,struct_plast_rev}. Several experimental studies have provided evidence for the formation and elimination of synapses in the adult brain \citep{syn_plast_imaging,spine_growth}, including activity-dependent pruning of weaker synapses during early development \citep{syn_pruning}. Large-scale growth and remodeling of axonal and dendritic branches has been found to occur in adult brain within days \citep{axon_growth}. In vivo imaging studies have shown that synaptic rewiring mediated by re-routing of whole axonal branches to different postsynaptic targets takes place in the mature cortex \citep{axon_rerout}.

Based on the above studies, we can conclude that structural changes provide an alternative form of plasticity in addition to synaptic weight changes associated with a classical Hebbian scheme. However, these two modes of long-term memory might operate on different timescales, where changes in synaptic efficacy could happen in seconds to hours while structural changes might operate over longer periods of days to months. The information storage capacity associated with structural plasticity lies in the ability to change wiring diagram in a sparsely connected network, which provides a large number of functionally distinct circuits available to encode information \citep{rewiring_chklo}.

\subsection{Comparisons with Related Work}

Some earlier studies have presented learning algorithms partly related to our work. \citet{fusi_mnist} used binary synapses and a stochastic spike-driven synaptic plasticity rule to train the network. Classification accuracy is obtained by a voting mechanism. The use of multiple output neurons for each class results in a high number of synapses. In comparison to this study, we use ``sparsely" connected binary synapses and dendritic nonlinearity to get high accuracy. The sparsity of synapses helps in learning as well as reduces the number of synapses used. 

Structural plasticity is used to modify binary synaptic connections on dendritic branches in \citet{mel_dendrite1}. The authors also use silent synapses to search for the best possible connection, which can replace a poorly performing active synapse. Compared to this study, our method of calculating $c_{ij}$ only needs values of inputs and dendritic branch outputs instead of derivatives as used by \citet{mel_dendrite1}, rendering our model more feasible for hardware implementation. In their study, the learning rule involved $g'()$ and $b'()$ which are derivatives of $g()$ and $b()$ functions respectively, where $g()$ was a sharp sigmoid and $b()$ was a degree $10$ polynomial, which is difficult to implement in hardware systems as compared to the quadratic function used in our work. We compared the performance of our learning rule with that in \citet{mel_dendrite1} without the $g'()$ term since it is non-zero only in a small region around origin, resulting in only marginally misclassified patterns with non-zero fitness values contributing to connection changes. The results (not shown in the paper) suggest that using either $b()$ in our model or $b'()$ in \citet{mel_dendrite1} does not lead to a significant difference in the classification performance. 

The unsupervised branch-specific STDP rule has been proposed by \citet{maass_bsp}, where the weight change of a synapse depends on the local dendritic branch potential. This was shown to allow a neuron to solve the binding problem. The supervised BSTDSP rule we proposed also uses branch-specific spike-time based correlations but in contrast to \citet{maass_bsp}, we use the correlation values to perform structural changes. 

\section{Conclusion and Future Work} 
\label{sec:conclusion}

This paper introduces two different learning schemes for spike-based representation in a nonlinear dendritic neuron model with the aim for implementation in neuromorphic hardware for which, we have (1) used a learning rule which is more amenable to hardware implementation; (2) tested the sensitivity of our models to noisy spike train inputs; (3) developed and tested a reduced model (RM) to minimize the training time and a margin-based learning method (RMWM) which led to a significant improvement in the noise sensitivity performance of our model; (4) proposed a branch-specific spike time dependent structural plasticity (BSTDSP) rule tying it to the well-known spike-based learning protocols; and (5) finally, compared the generalization performance of our model with state-of-the-art nonlinear classifiers on real-world classification tasks and showed that our neurally-inspired model gives similar performance to these other methods. 

In the future, we shall extend our spike-based algorithm which combines both structural learning with STDP, to incorporate precise spike timing information for learning both synaptic connections as well as the weights of these synapses on a dendritic branch. Also, we shall incorporate the effect of a specific spatial location of a synapse on the same dendritic branch by including location dependent delays as an extra tuning parameter \citep{shaista_apccas,neurocomputing_delay}. We are also exploiting the silicon-friendly features of this algorithm to develop integrated circuits implementing the model described here for a spike-based smart sensor.

\section*{Acknowledgement}Financial support from MOE through grant ARC 8/13 is acknowledged.

\section*{Appendix}

\subsection{Hardware Implementation}

The classifier model presented is suited to VLSI implementation since it does not require high resolution weights; instead, it only requires the storage of a sparse connection matrix in memory. For applications where the statistics of the input are known, a connectivity matrix obtained through offline training can be downloaded to the hardware. However, input statistics may not be known a priori in many cases, and the connections might have to be learned in the field. The $c_{ij}$ values guiding this connection change will thus need to be computed on-chip and the current learning rule is designed to facilitate that. The dendritic output ($b_j$) can be computed in current mode using a translinear loop as shown in \citet{ijcnn_dendrite}--only one such circuit is needed per dendrite. Each synapse, implemented using a spike to current conversion circuit \citep{barto_indi_synapse}, needs to have an associated current mode multiplier to compute the product $x_{ij}b_j$, where $x_{ij}$ is the current output of the synapse. The last part of the $c_{ij}$ term can be implemented using switches controlled by digital circuits since $sgn(o-y)$ only takes $3$ discrete values and both $o$ and $y$ are digital signals. To encode the $3$ values of the $sgn$ function, we need two digital wires such that $01$, $10$ and $00$ represent the values $-1$, $1$ and $0$ respectively. These two wires can control two separate switches diverting the current encoding $x_{ij}b_j$ to a positive or a negative integrator for the $01$ and $10$ codes. Both switches are off for the $00$ code thus not adding to the integral.

\subsection{Validity of the Reduced Model}
\label{sec:validity}

The RM and RMWM employ time-averaged synaptic activation, $z_{syn,ij}$ instead of the actual postsynaptic current generated by a synapse. This approximation can be used to describe the average current for the $j^{th}$ dendrite as in the following equation:
\begin{equation}
\tilde{I}^j_{b,out}=\left\{ \sum_{i}w_{ij}(\frac{1}{T}\int_{0}^{T}(\sum_{t^s_{ij}}K(t-t^s_{ij}))dt) \right\}^2 / x_{thr}
\end{equation}
where the dendritic nonlinearity is a quadratic function. However, the actual dendritic current is given by the following equation:
\begin{equation}
I^j_{b,out}(t)=\left\{ \sum_{i}w_{ij}\sum_{t^s_{ij}}K(t-t^s_{ij}) \right\}^2 / x_{thr}
\end{equation}

Therefore, the reduced models are valid if the average dendritic current $\tilde{I}^j_{b,out}$ is equal to the average current computed using the spike-based model, i.e.
\begin{align}
\label{eq:act_red}
\tilde{I}^j_{b,out}&=\frac{1}{T}\int_{0}^{T}I^j_{b,out}(t)dt \notag\\
\Rightarrow \left\{ \sum_{i}w_{ij} (\frac{1}{T}\int_{0}^{T} (\sum_{t^s_{ij}}K(t-t^s_{ij}))dt) \right\}^2&=\frac{1}{T}\int_{0}^{T}\left\{ \sum_{i}w_{ij}\sum_{t^s_{ij}}K(t-t^s_{ij}) \right\}^2dt
\end{align}

\begin{figure}[!t]
\centerline{
\includegraphics[width=0.6\textwidth]{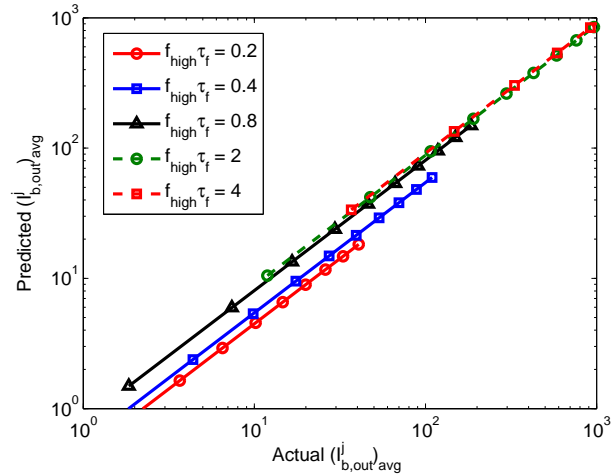}}
\caption{Average dendritic current predicted using the reduced model (y-axis) plotted against actual mean current (x-axis) for different values of $f_{high} \times \tau_f$ and different number of synapses on a dendrite.}
\label{fig:Iact_red}
\end{figure}

The equation \eqref{eq:act_red} will hold true only if the postsynaptic current $\sum_{t^s_{ij}}K(t-t^s_{ij})$ is a constant function of time. For the PSC waveform to be described by a constant function, both the frequency of a spike train arriving at a synapse ($f_{high}$) and synaptic current fall time constant ($\tau_f$ in equation \eqref{eq:kernel}) are increased. Figure \ref{fig:Iact_red} shows that as $f_{high} \times \tau_f$ increases, the plot of average dendritic current calculated using the reduced models (y-axis) versus the actual values (x-axis) drifts toward the `y=x' line, which indicates that the values predicted by the reduced model become equal to the actual values. The data points correspond to different number of synapses on a branch. These results have guided the choice of $f_{high}$ and $\tau_f$ in this work.

\bibliographystyle{chicago}

\bibliography{references}

\end{document}